\documentclass[10pt,twocolumn,letterpaper]{article}

\usepackage[pagenumbers]{cvpr}      

\usepackage{graphicx}
\usepackage{amsmath}
\usepackage{amssymb}
\usepackage{booktabs}
\usepackage{multirow}
\usepackage{tabularx}
\usepackage{float}
\usepackage[accsupp]{axessibility}
\usepackage{setspace}

\usepackage[pagebackref,breaklinks,colorlinks]{hyperref}

\usepackage[capitalize]{cleveref}
\crefname{section}{Sec.}{Secs.}
\Crefname{section}{Section}{Sections}
\Crefname{table}{Table}{Tables}
\crefname{table}{Tab.}{Tabs.}

\def\cvprPaperID{1538} 
\def\confName{CVPR}
\def\confYear{2023}

\begin{document}

\title{PiMAE: Point Cloud and Image Interactive \\Masked Autoencoders for 3D Object Detection}



\author{Anthony Chen$^{1,2,*}$, Kevin Zhang$^{1,2,*}$, Renrui Zhang$^{3}$, Zihan Wang$^{1,2}$, \\ Yuheng Lu$^{1,2,4}$, Yandong Guo$^{5}$, Shanghang Zhang$^{1,2,\dagger}$ \\
$^1$National Key Laboratory for Multimedia Information Processing \\
$^2$Peking University \ \ $^3$The Chinese University of Hong Kong \\
$^4$Wukong Lab, iKingtec \ \ $^5$Beijing University of Posts and Telecommunications\\ 
\texttt{\{antonchen, kevinzyz6, shanghang\}@pku.edu.cn} \\
\texttt{zhangrenrui@pjlab.org.cn},\ \ 
\texttt{rzhevcherkasy@outlook.com} \\
\texttt{luyuheng@ikingtec.com}, \ \ 
\texttt{yandong.guo@live.com}
}
\maketitle

\renewcommand{\thefootnote}{\fnsymbol{footnote}} 
\footnotetext[1]{Equal Contribution.} 
\footnotetext[2]{Corresponding Author.} 



\begin{abstract}

    \vspace{-0.035in}
   Masked Autoencoders learn strong visual representations and achieve state-of-the-art results in several independent modalities,
   yet very few works have addressed their capabilities in multi-modality settings. In this work, we focus on point cloud and RGB image data, two modalities that are often presented together in the real world, and explore their meaningful interactions.
   To improve upon the cross-modal synergy in existing works, we propose PiMAE, a self-supervised pre-training framework that promotes 3D and 2D interaction through three aspects.
   Specifically, we first notice the importance of masking strategies between the two sources and utilize a projection module to complementarily align the mask and visible tokens of the two modalities. Then, we utilize a well-crafted two-branch MAE pipeline with a novel shared decoder to promote cross-modality interaction in the mask tokens. Finally, 
   we design a unique cross-modal reconstruction module to enhance representation learning for both modalities. Through extensive experiments performed on large-scale RGB-D scene understanding benchmarks (SUN RGB-D and ScannetV2), we discover it is nontrivial to interactively learn point-image features, where we greatly improve multiple 3D detectors, 2D detectors, and few-shot classifiers by 2.9\%, 6.7\%, and 2.4\%, respectively. Code is available at \textcolor{magenta}{https://github.com/BLVLab/PiMAE}.
\end{abstract}


\begin{figure} 
    \centering 
    \includegraphics[width=0.47\textwidth]{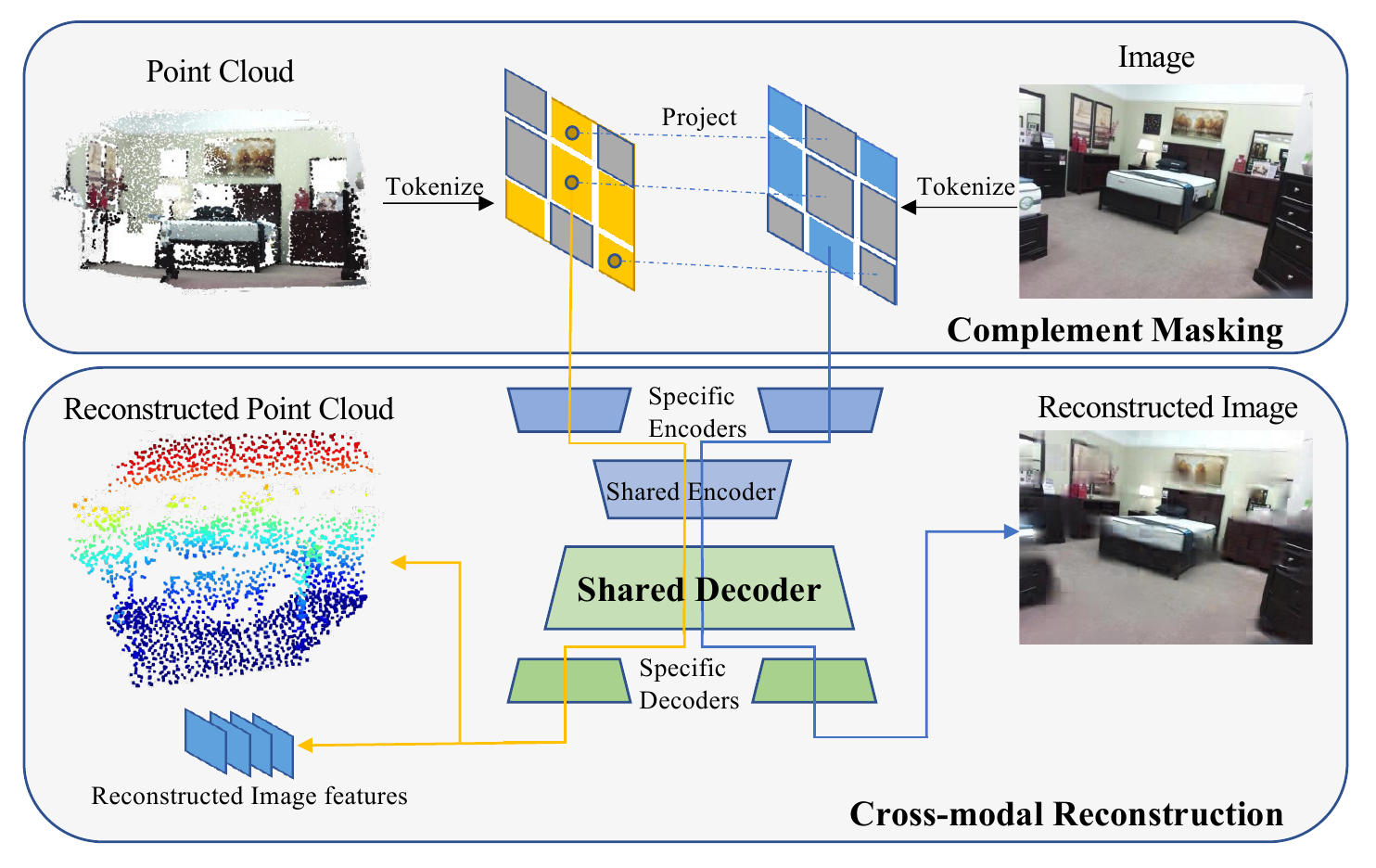}
    \caption{With our proposed design, PiMAE learns cross-modal representations by interactively dealing with multi-modal data and performing reconstruction.} 
    \label{Fig.main2} 
\end{figure}

\section{Introduction}

\label{sec:intro}

The advancements in deep learning-based technology have developed many significant real-world applications, such as robotics and autonomous driving. In these scenarios, 3D and 2D data in the form of point cloud and RGB images from a specific view are readily available. Therefore, many existing methods perform multi-modal visual learning, a popular approach leveraging both 3D and 2D information for better representational abilities.

Intuitively, the paired 2D pixels and 3D points present different perspectives of the same scene. They encode different degrees of information that, when combined, may become a source of performance improvement.
Designing a model that interacts with both modalities, such as geometry and RGB, is a difficult task because directly feeding them to a model results in marginal, if not degraded, performance, as demonstrated by~\cite{P4Constrast }.

In this paper, we aim to answer the question: 
how to design a more interactive unsupervised multi-modal learning framework that is for better representation learning?
To this end, we investigate the Masked Autoencoders (MAE) proposed by He \etal~\cite{he2022masked}, which demonstrate a straightforward yet powerful pre-training framework for Vision Transformers\cite{vit} (ViTs) and show promising results for independent modalities of both 2D and 3D vision \cite{convmae, bachmann2022multimae, M3AE, pointm2ae, zhang2022i-mae}.  
However, these existing MAE pre-training objectives are limited to only a single modality.

While much literature has impressively demonstrated MAE approaches' superiority in multiple modalities, existing methods have yet to show promising results in bridging 3D and 2D data. For 2D scene understanding among multiple modalities, MultiMAE~\cite{bachmann2022multimae} generates pseudo-modalities to promote synergy for extrapolating features. Unfortunately, these methods rely on an adjunct model for generating pseudo-modalities, which is sub-optimal and makes it hard to investigate cross-modality interaction. 
On the other hand, contrastive methods for self-supervised 3D and 2D representation learning, such as \cite{xie2020pointcontrast, chen2021mocov3, chen2020simple, P4Constrast,crosspoint}, suffer from sampling bias when generating negative samples and augmentation, making them impractical in real-world scenarios~\cite{NEURIPS2020_63c3ddcc,zhao2021graph,hong2021unbiased}. 





To address the fusion of multi-modal point cloud and image data, we propose PiMAE, a simple yet effective pipeline that learns strong 3D and 2D features by increasing their interaction. Specifically, we pre-train pairs of points and images as inputs, employing a two-branched MAE learning framework to individually learn embeddings for the two modalities. To further promote feature alignment, we design three main features.



First, we tokenize the image and point inputs, and to correlate the tokens from different modalities, we project point tokens to image patches, explicitly aligning the masking relationship between them. We believe a specialized masking strategy may help point cloud tokens embed information from the image, and vice versa.
Next, we utilize a novel symmetrical autoencoder scheme that promotes strong feature fusion. The encoder draws inspiration from \cite{AIST}, consisting of both separate branches of modal-specific encoders and a shared-encoder. However, we notice that since MAE's mask tokens only pass through the decoder\cite{he2022masked}, a shared-decoder design is critical in our scheme for mask tokens to learn mutual information before performing reconstructions in separate modal-specific decoders.
Finally, for learning stronger features inspired by~\cite{zhang2022learning,gaomimic}, PiMAE's multi-modal reconstruction module tasks point cloud features to explicitly encode image-level understanding through enhanced learning from image features.

To evaluate the effectiveness of our pre-training scheme, we systematically evaluate PiMAE with different fine-tuning architectures and tasks, including 3D and 2D object detection and few-shot image classification, performed on the RGB-D scene dataset SUN RGB-D~\cite{sunrgbd} and ScannetV2 \cite{dai2017scannet} as well as multiple 2D detection and classification datasets. We find PiMAE to bring improvements over state-of-the-art methods in all evaluated downstream tasks.

Our main contributions are summarized as:
\begin{itemize}
    \item To the best of our knowledge, we are the first to propose pre-training MAE with point cloud and RGB modalities interactively with three novel schemes.
    %
    \item To promote more interactive multi-modal learning, we novelly introduce a complementary cross-modal masking strategy, a shared-decoder, and cross-modal reconstruction to PiMAE.
    \item Shown by extensive experiments, our pre-trained models boost performance of 2D \& 3D detectors by a large margin, demonstrating PiMAE's effectiveness.
\end{itemize}




\section{Related Work}
\label{sec:formatting}

\textbf{3D Object Detectors.}
3D object detection aims to predict oriented 3D bounding boxes of physical objects from 3D input data. Many CNN-based works propose two-stage methods for first generating region proposals and then classifying them into different object types.
Prior 3D object detection methods adapt popular 2D detection approaches to 3D scenes, projecting point cloud data to 2D views \cite{conv1, conv2, Xu_2018_CVPR} for 2D ConvNets to detect 3D bounding boxes. Other approaches adopt 3D ConvNets by grouping points into voxels \cite{voxelnet, imvoxelnet} and transposed convolutions for sparse detection~\cite{gwak2020gsdn}.
Recently, the Transformer architecture~\cite{transformers} has demonstrated consistent and impressive performance in vision, specifically with object detectors \cite{detr, Liu2020TANetR3,pointTrans, mao2021dual, zheng2020end, groupfree,misra2021-3detr, shi20193d,shi2020pv,rukhovich2021fcaf3d,zhang2023parameter}. Transformers are especially well-designed for 3D point clouds, needing not hand-crafted groupings and capable of having an invariant understanding. Specifically, ~\cite{misra2021-3detr} proposed an end-to-end Transformer-based object detection module using points cloud as input. Group-Free-3D~\cite{groupfree} designed a novel attention stacking scheme and estimated detection results by fusing object features in different stages.
In PiMAE, we draw inspiration from both projection-based and attention-based 3D object detectors; whereas the former projection mechanisms have been extensively utilized previously, the latter has shown better versatility and a more intuitive solution. Consequently, we design a MAE~\cite{he2022masked}-structured multi-modal learning framework that incorporates projection alignment for more interactive multi-modal learning.

\textbf{Point Cloud and Image Joint 
Representation Learning.}
3D point cloud and 2D image joint representation learning methods aim to explore the modal interaction between point clouds and images for feature fusion. Many recent studies have shown that cross-modal modules outperform single-modal methods on multiple tasks such as 3D object 
detection~\cite{pri3d, H3DNet,yang2022boosting,wang2021pointaugmenting,huang2022tig,wu2022eda}, 3D semantic segmentation~\cite{3D-SIS, peng2021sparse,jaritz2022cross}, and 3D open-world learning~\cite{zhu2022pointclip,zhang2022pointclip,zhang2022can,guo2022calip}.
In cross-modal self-supervised learning of point clouds and RGB images, several methods~\cite{li2022simipu, P4Constrast, houji} based on contrastive learning propose to design specialized structures for learning from multiple modalities surpass single modalities when fine-tuned on downstream tasks including 3D object detection.
As aforementioned, while contrastive methods have illustrated the significance of pairing RGB and point clouds, PiMAE has several advantages over contrastive methods, mainly requiring fewer augmentations.

\textbf{Masked Autoencoders (MAE).}
Recently, inspired by advances in masked language modeling, masked image modeling (MIM) approaches~\cite{he2022masked, xie2021simmim, beit} have shown superior performance, proposing a self-supervised training method based on masked image prediction. MAE~\cite{he2022masked}, in particular, predicts pixels from highly masked images using a ViT decoder.
Since MAE's success, several works~\cite{pointm2ae,pointmae,min2022voxel,zhang2021self,hess2022masked,fu2022pos} have applied the framework to point cloud data, proposing to segment point cloud into tokens and perform reconstruction. Moreover, MultiMAE~\cite{bachmann2022multimae} investigates the alignment of various modalities with MAE among RGB images, depth images, and semantic segmentation. Recently, I2P-MAE~\cite{zhang2022learning} explores leveraging 2D pre-trained knowledge to guide 3D MAE pre-training. In this work, however, we demonstrate that earlier methods do not maximize the potential of point cloud and RGB scene datasets, because they cannot incorporate the RGB inputs with ease and bring only trivial performance gain. To the best of our knowledge, this is the pioneering work aligning RGB images with point cloud with MAE pre-training.
\section{Methods}
\label{sec:methods}

\begin{figure} [t]
    \centering 
    \includegraphics[width=0.4\textwidth]{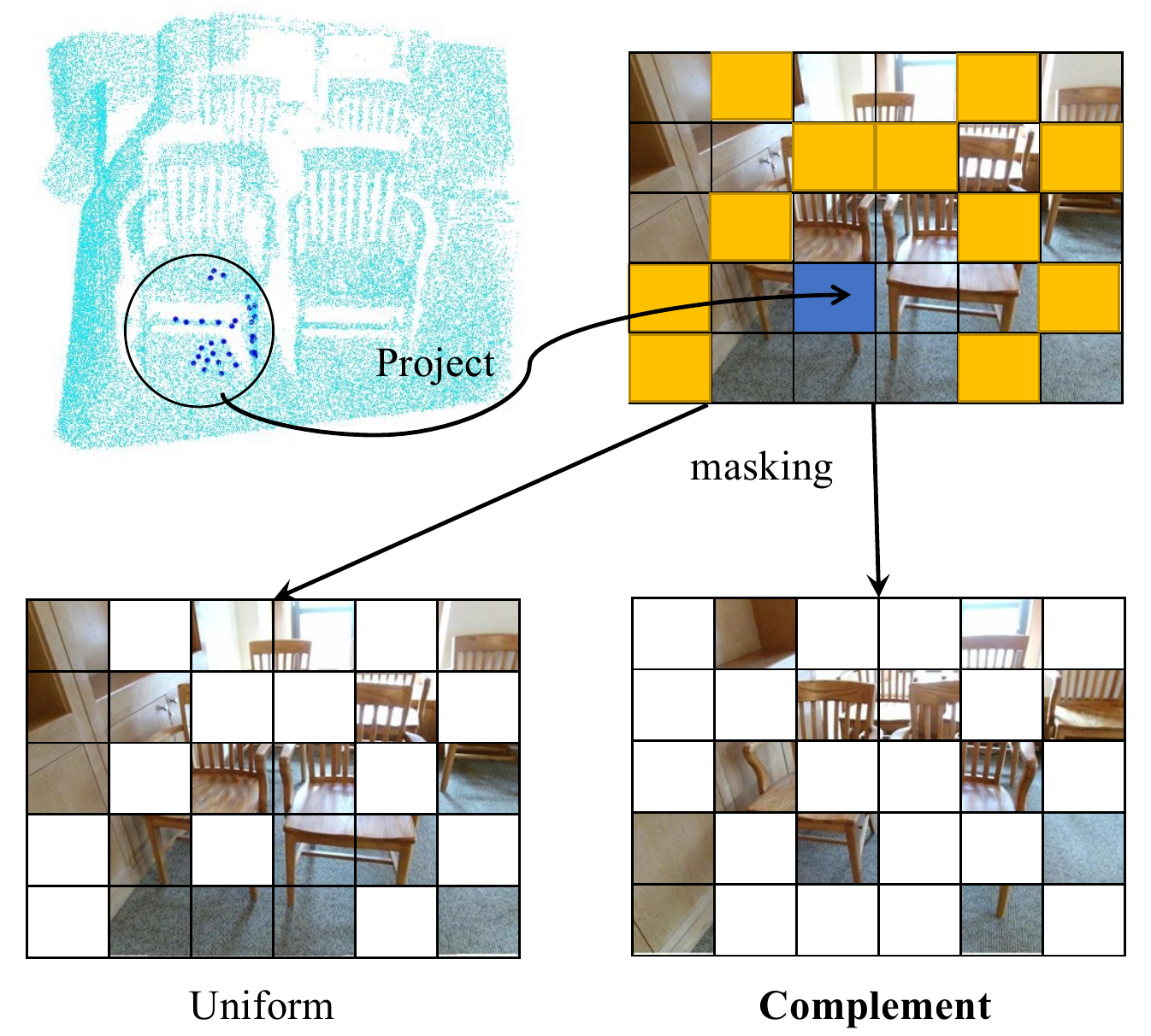} 
    \vspace{-0.05in}
    \caption{\textbf{Illustration of our projection operation and two different masking strategies.} A randomly sampled point cloud cluster (black circle) is projected onto the image patch (blue square), and the other clusters are done in a similar way (yellow squares). Under uniform masking, the yellow patches will be masked while other patches are visible. On the contrary, complement masking will result in a reversed masking pattern.} 
    \label{fig:masking} 
\end{figure}

\begin{figure*} [t]
  \centering
    \includegraphics[width=0.9\linewidth]{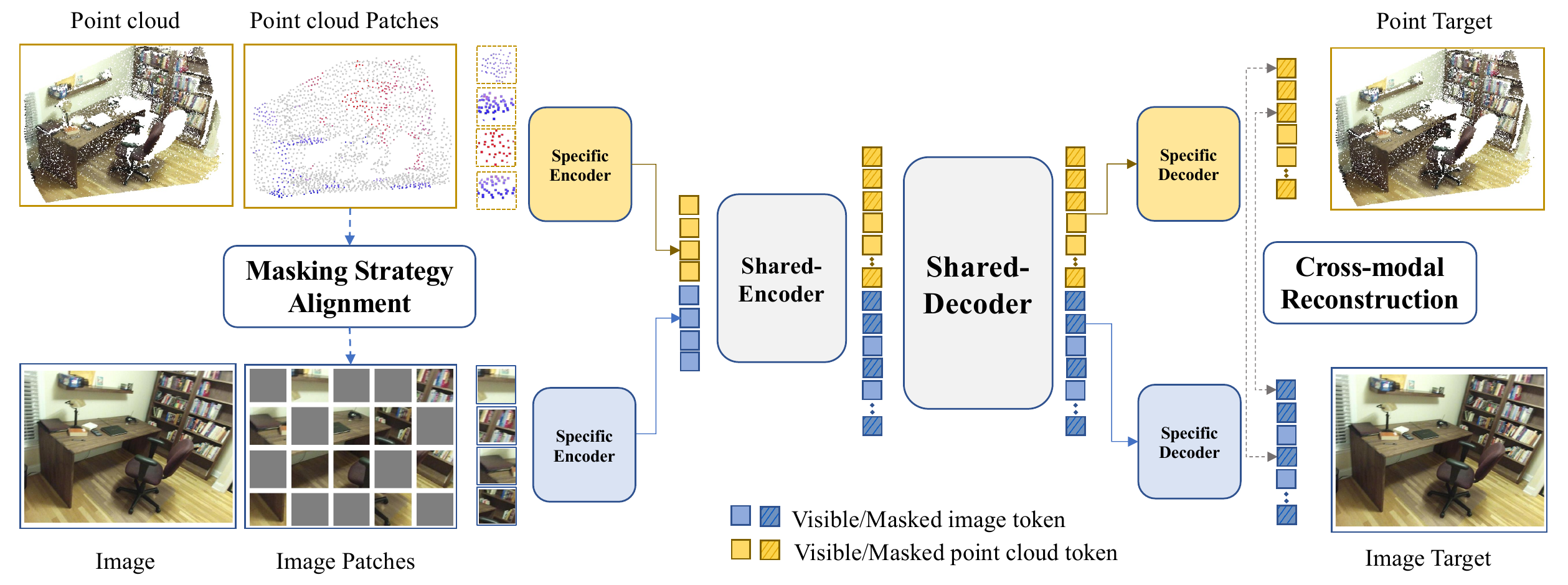}
  \caption{\textbf{Pre-training pipeline for PiMAE.} The point cloud branch samples and clusters point cloud data into tokens and randomly masks the input. 
  Then the tokens pass through the masking alignment module to generate complement masks for image patches. After embedding, tokens go through a separate, then shared, and finally separated autoencoder structure. We lastly engage in a cross-modal reconstruction module to enhance point cloud representation learning. Point cloud is colored for better visualization.}
  \label{fig:pipeline}
\end{figure*}

In this section, we first give an overview of our pipeline. Then, we introduce our novelly 
designed masking strategy, which aligns the semantic information between tokens from two modalities. Followingly, we present our cross-modal encoders and decoders design. Notably, the shared-decoder is a pioneering architecture. Finally, we finish with our cross-modal reconstruction module.

\subsection{Pipeline Overview}

As shown in Fig.~\ref{fig:pipeline}, PiMAE learns cross-modal representations simultaneously by jointly learning features from point clouds and image modalities. In our proposed pipeline, we first embed point data into tokens by sampling and clustering algorithms and then perform random masking on point tokens. The mask pattern is transformed onto the 2D plane, where patches of images are complementarily masked and embedded into tokens.  

Following this, we utilize a symmetrical joint-encoder-decoder scheme that promotes strong feature fusion. The encoder-decoder architecture consists of both separate branches and shared modules, whereas the former protects modal-specific learning and the latter encourages cross-modal interaction for more robust features.

Finally, for learning stronger features from pre-training, PiMAE's cross-modal reconstruction module demands point cloud features to explicitly express image-level understanding. 

\subsection{Token Projection and Alignment}

We follow MAE\cite{he2022masked} and Point-M2AE\cite{pointm2ae} to generate input tokens from images and point clouds. An image is first divided into non-overlapping patches, before the embedding procedure that embeds patches by a linear projection layer with added Positional Embeddings (PE) and Modality Embeddings (ME). Correspondingly, a set of point clouds is processed into cluster tokens via Farthest Point Sampling (FPS) and K-Nearest Neighbour (KNN) algorithms and then embedded with a linear projection layer with added embeddings (i.e. PE, ME).

\textbf{Projection.} 
\label{sec:proj}
In order to achieve the alignment between multi-modality tokens, we build a link between the 3D point cloud and RGB image pixels by projecting the point cloud onto the camera's image plane. For 3D point $P \in \mathbb{R}^3$, a correlating 2D coordinate can be calculated using the projection function $Proj$ defined below,
\begin{equation}
    \label{eq:projection}
    \begin{bmatrix} u \\ v \\z \end{bmatrix} = Proj(P) = K \cdot R_t \cdot \begin{bmatrix} x \\ y \\ z \\ 1 \end{bmatrix}, 
\end{equation}
where $K \in 3\times 4$, $R_t \in 4\times 4$ are the camera intrinsic and extrinsic matrices. $(x,y,z)$, $(u,v)$ are the original 3D coordinate and projected 2D coordinate of point $P$.

\textbf{Masking with Alignment.} 
\label{sec:masking}
Next, we generate masked tokens using the aforementioned projection function. Since point cloud tokens are organized by the cluster centers, we randomly select a subset of center points as well as their corresponding tokens, while keeping the rest masked. 
For the visible point cloud tokens $T_p$, we project their center point $P \in \mathbb{R}^3$ to the corresponding camera plane and attain its 2D coordinate $p \in \mathbb{R}^2$, which can naturally fall into an area of shape $H \times W$ (i.e. image shape), thus obtaining its related image patch index $I_p$ by

\begin{equation}
    \label{eq:indexing}
    I_p = \lfloor \lfloor v \rfloor / S \rfloor \times \lfloor W / S \rfloor + \lfloor \lfloor u \rfloor / S \rfloor,
\end{equation}
where $u$ and $v$ denotes the $x$-axis value and $y$-axis value of 2D coordinate $p$, $S$ is the image patch size.

After projecting and indexing each visible point cloud token, we obtain their corresponding image patches. Next, we explicitly mask these patches to reach a complement mask alignment. The rationale is that such masking pattern can make visible tokens more semantically abundant than under the uniform setting, and thus the model is able to extract rich cross-modal features. For visual demonstration of our projection and alignment, see Fig.~\ref{fig:masking}.

\subsection{Encoding Phase}

\textbf{Encoder.}
\label{sec:encoder}
During this stage, we protect the integrity of different modalities. Inspired by AIST++~\cite{AIST}, our encoder consists of two modules: modal-specific encoders and a cross-modal encoder. The former is used to better extract modal-specific features, and the latter is used to perform interaction between cross-modal features. 

The modality-specific encoder part contains two branches for two modalities, where each branch consists of a ViT backbone.
First, for the encoders to learn modality differences through mapping inputs to feature spaces, we feed the aligned, visible tokens with their respective positional and modality embeddings to separate encoders. 

Later, we promote feature fusion and cross-modality interactions of visible patches with a shared-encoder. The alignment of masks during this stage becomes critical, as aligned tokens reveal similar information reflected in both 3D and 2D data.

Formally, in the separate encoding phase, $E_I: T_I\mapsto L_I^1$ and $E_P: T_P\mapsto L_P^1$, where $E_I$ and $E_P$ are the image-specific and point-specific encoders, $T_I$ and $T_P$ are the visible image and point patch tokens, and $L_I^1$ and $L_P^1$ are the image and point latent spaces. Then, the shared-encoder performs fusion on the different latent representations $E_S: L^1_I, L^1_P \mapsto L^2_S$.

\subsection{Decoding Phase}

\textbf{Decoder.}
\label{sec:decoder}
Generally, MAE encoders benefit from learning generalized encoders that capture high-dimensional data encoding representations for both image and point cloud data. Due to the differences between the two modalities, specialized decoders are needed to decode the high-level latent to the respective modality.

The purpose of additional shared-decoder layers is ultimately for the encoder to focus more on feature extraction and ignore the details of modality interactions.
Because MAE uses an asymmetric autoencoder design where the mask tokens do not pass the shared-encoder, we complement the mask tokens to pass through a shared-decoder, along with the visible tokens. 
Without such a design, the entire decoder branches are segmented, and the mask tokens of different modalities do not engage in feature fusion. After the shared-decoder, we then design specialized decoders for the different modalities for better reconstructions. 

Since the reconstructions of two modalities are involved, the losses of both the point cloud and the image modalities are obtained. For point clouds, we use $\ell_2$ Chamfer Distance \cite{chamferdistance} for loss calculation, denoted as $\mathcal{L}_{pc}$, and for images, we use MSE to measure the loss, denoted as $\mathcal{L}_{img}$. 

Formally, the input for our shared-decoder is $L_S^{2'}$, the full sets of tokens including encoded visible features and mask tokens of both modalities, and the shared-decoder performs cross-modal interaction on these latent representations $D_S: L_S^{2'} \mapsto L_I^3, L_I^3$.
Then, in the separate decoder phase, the decoders map back to the image and point cloud space, $D_I: L_I^3\mapsto T^{'}_I$ and $D_P: L_P^3\mapsto T^{'}_P$, where $D_I$ and $D_P$ are the image-specific and point-specific decoders, $T^{'}_I$ and $T^{'}_P$ are the visible image and point cloud patches, and $L_I^3$ and $L_P^3$ are the image and point cloud latent spaces.

\begin{equation}
         \mathcal{L}_{pc} = CD(D_P(l), P_{GT}),
\end{equation}

where $CD$ is $\ell_2$ Chamfer Distance function \cite{chamferdistance}, $D_P$ represents the decoder reconstruction function, 
$l \in L_P^3$ is the point cloud latent representation,
$P_{GT}$ is the ground truth point cloud (i.e. point cloud input). 

\subsection{Cross-modal Reconstruction}
\label{sec:loss}

We train PiMAE using three different losses: the point cloud reconstruction loss, the image reconstruction loss, and a cross-modal reconstruction loss that we design to further strengthen the interaction between the two modalities. 
In the final reconstruction phase, we utilize the previously aligned relationship to obtain the corresponding 2D coordinates of the masked point clouds.
Then, we up-sample the reconstructed image features, such that each masked point cloud with a 2D coordinate can relate to a reconstructed image feature. Finally, the masked point cloud tokens go through a cross-modal prediction head of one linear projection layer to recover the corresponding visible image features. Note that we specifically avoid using visible point cloud tokens for this module, because they correspond to the masked image features (due to the complement masking strategy), which tend to have weaker representations and may harm representation learning. Formally, the cross-modal reconstruction loss is defined as

\begin{equation}
     \mathcal{L}_{cross} = MSE(D_P(l_p^3), l_i^3),
\end{equation}

where $MSE$ denotes the Mean Squared Error loss function, $D_P$ is the cross-modal reconstruction from the decoder, 
$l_p^3 \in L_P^3$ is the point cloud representation,
$l_i^3 \in L_I^3$ is the image latent representation.

Our final loss is the sum of the previous loss terms, formulated in Eq.~\ref{eq:finalloss}. By such design, PiMAE learns 3D and 2D features separately while maintaining strong interactions between the two modalities.

\begin{equation}
     \mathcal{L} = \mathcal{L}_{pc} + \mathcal{L}_{img} +  \mathcal{L}_{cross}.
    \label{eq:finalloss}
\end{equation}

\begin{figure*}
  \centering
  \begin{subfigure}{\linewidth}
  \centering
    {\includegraphics[width=\linewidth]{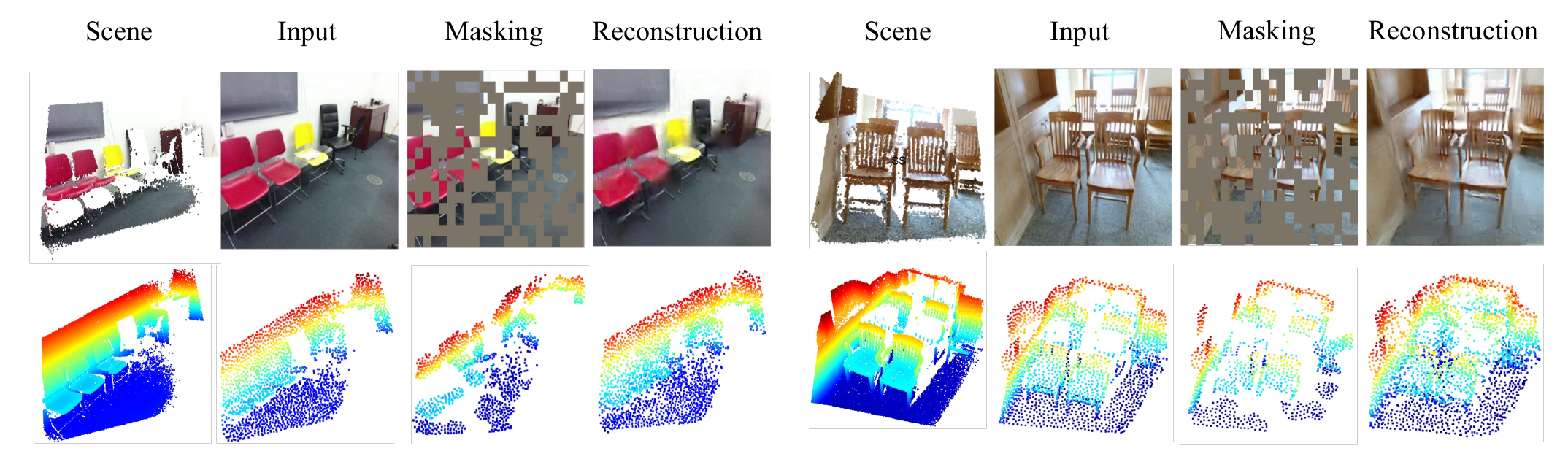}
    }
    \label{fig:short-a}
  \end{subfigure}
  \hfill
  \vspace{-0.3in}
  \caption{\textbf{Reconstruction results of images and point cloud from PiMAE.} Our model is able to perform image and point cloud reconstruction simultaneously, showing a firm understanding of the two modalities. Image results are in the first row, point cloud results in the second. The masking ratio for both branches is 60\%. Point cloud is colored for better visualization.}
  \label{fig:reconstruction}
\end{figure*}

\section{Experiments}
\label{sec:experiments}

\begin{table*}[h]
\centering

\begin{tabular}{l|ccccc}
\toprule
 & & \multicolumn{2}{c}{SUN RGB-D} & \multicolumn{2}{c}{ScanNetV2} \cr
Methods & Pre-trained & $AP_{25}$ & $AP_{50}$ &  $AP_{25}$ & $AP_{50}$\\
\midrule
DSS\cite{DSS} & \textit{None} & 42.1 & - & 15.2 & 6.8 \\
PointFusion\cite{PointFusion} & \textit{None} & 45.4 & - & - & - \\
3D-SIS\cite{3D-SIS} & \textit{None} & - & - & 40.2 & 22.5 \\
VoteNet\cite{votenet} & \textit{None} & 57.7 & 32.9 & 58.6 & 33.5 \\
\midrule
 3DETR\cite{misra2021-3detr} & \textit{None} & 58.0 & 30.3 & 62.1 & 37.9 \\
 +Ours(from scratch) & \textit{None} & 58.7 & 31.7 & 59.7 & 40.0 \\
 +Ours & SUN RGB-D & 59.4(+1.4) & 33.2(+2.9) & 62.6(+0.5) & 39.4(+1.5) \\
\midrule
GroupFree3D\cite{groupfree} & \textit{None} & 63.0 & 45.2 & 67.3 & 48.9 \\
 +Ours(from scratch) & \textit{None} & 61.2 & 44.7 & 65.5 & 47.4 \\
 +Ours & SUN RGB-D & \bf 64.6(+1.6) & \bf 46.2(+1.0) & \bf 67.6(+0.3) & \bf 49.7(+0.8) \\ 
\bottomrule
\end{tabular}
\caption{\textbf{3D object detection results on ScanNetV2\cite{dai2017scannet} and SUN RGB-D\cite{sunrgbd}.} We adopt the average precision with 3D IoU thresholds of 0.25 ($AP_{25}$) and 0.5  ($AP_{50}$) for the evaluation metrics.}
\label{tab:detection}
\end{table*}

\begin{table*} [t]
\begin{minipage}[t]{0.48\textwidth}
\makeatletter\def\@captype{table}
\begin{tabular}{lccc} 
    \toprule
    Methods & $AP_{50}$ & $AP_{75}$ & $AP$ \\
    \midrule
    *DETR\cite{detr} & 39.8 & 26.2 & 25.3 \\ 
     + PiMAE & \textbf{46.5(+6.7)} & \textbf{30.3(+4.1)} & \textbf{29.5(+4.2)} \\ 
    \bottomrule
  \end{tabular}
  \vspace{-0.05in}
\caption{\textbf{2D object detection results on ScanNetV2 val set.} \\ * denotes our implementation on ScanNetV2. We later load \\ PiMAE pre-trained weights to the encoder.}
\label{tab:2ddection}
\end{minipage}
\begin{minipage}[t]{0.48\textwidth}
\makeatletter\def\@captype{table}
\begin{tabular}{lccc} 
    \toprule
     Methods & Easy & Mod. & Hard \\
    \midrule
    *MonoDETR\cite{zhang2022monodetr} & 23.1 & 17.3 & 14.5 \\ 
     + PiMAE & \textbf{26.6(+3.5)} & \textbf{18.8(+1.5)} & \textbf{15.5(+1.0)} \\ 
    \bottomrule
  \end{tabular}
  \vspace{-0.05in}
\caption{\textbf{Monocular 3D object detection results of car category on KITTI \textit{val} set.} * denotes our implementation with adjusted depth encoder, to which we later load PiMAE pre-trained weights.}
\label{tab:mono3d}
\end{minipage}
\end{table*}

In this section, we provide extensive experiments to qualify the superiority of our methods. The following experiments are conducted. a) We pre-train PiMAE on the SUN RGB-D~\cite{sunrgbd} training set. b) We evaluate PiMAE on various downstream tasks, including 3D object detection, 3D monocular detection, 2D detection, and classification. c) We ablate PiMAE with different multi-modal interaction strategies to show the effectiveness of our proposed design.

\subsection{Implementation Details}
\textbf{Datasets and metrics.} We pre-train our model on SUN RGB-D\cite{sunrgbd} and evaluate with different downstream tasks on several datasets including indoor 3D datasets (SUN RGB-D\cite{sunrgbd}, ScanNetV2\cite{dai2017scannet}), outdoor 3D dataset (KITTI\cite{kitti}), few-shot image classification datasets (CIFAR-FS\cite{cifar-fs}, FC100\cite{FC100}, miniImageNet\cite{miniImageNet}). Detailed descriptions of these datasets and evaluation metrics are in the Appendix.

\textbf{Network architectures.} Abiding by common practice~\cite{pointmae, pointm2ae}, we utilize a scaled-down PointNet~\cite{qi2016pointnet} before a ViT~\cite{vit} backbone in our point cloud branch. PointNet layers effectively reduce the sub-sampled points from 20,000 to 2,048. For the image branch, we follow~\cite{he2022masked} to divide images into regular patches with a size of $16 \times 16$, before the ViT backbone. 

\textbf{Pre-training}.
During this stage, we use the provided image and generated point cloud from SUN RGB-D~\cite{sunrgbd} to train PiMAE for 400 epochs. AdamW~\cite{adamw} optimizer with a base learning rate of 1e-3 and weight decay of 0.05 is used, applied with a warm-up for 15 epochs. No augmentation is performed on both image and point cloud inputs, for the main goal of maintaining consistency between the patches.
Experimentally, we find that a masking ratio of 60\% is more appropriate. The reconstructed visualization results are in Fig.~\ref{fig:reconstruction}.
Detailed configurations are in the Appendix.

\textbf{Fine-tuning.} 
With PiMAE's two multi-modal branches, we fine-tune and evaluate our learned features on both 3D and 2D tasks. 
For 3D tasks, we use the point cloud branch's specific encoder and the shared encoder as a 3D feature extractor. For 2D tasks, similarly, we utilize the image-specific encoder as well as the shared encoder as a 2D feature extractor. We fit our feature extractors into different baselines and keep the same training settings, except for the modifications on the backbone feature extractor. 
Detailed configurations are in the Appendix.

\subsection{Results on Downstream Tasks}
In this work, we evaluate our method on four different downstream tasks dealing with different modalities
, including 3D object detection, monocular 3D object detection, 2d object detection, and few-shot image classification.

\textbf{Indoor 3D object detection.}
We apply our 3D feature extractor on 3D detectors by replacing or inserting the encoder into different backbones to strengthen feature extraction. We report our performance on indoor 3D detection based on SOTA methods 3DETR\cite{misra2021-3detr} and GroupFree3D\cite{groupfree}. As shown in Tab.~\ref{tab:detection}, our model brings significant improvements to both models, surpassing previous baselines consistently in all datasets and criteria.

Furthermore, in the Appendix, we provide 3D object detection results with detailed per-class accuracy on SUN RGB-D, along with visualizations of detection results. 

\textbf{Outdoor monocular 3D object detection.}
To fully demonstrate the capacity of our approach, we report our performance in challenging outdoor scenarios, which have a large data distribution gap compared with our indoor pre-training data. As shown in Tab.~\ref{tab:mono3d}, we brought substantial improvement to MonoDETR~\cite{zhang2022monodetr}, validating that our pre-trained representations generalize well to both indoor and outdoor datasets.

\textbf{2D object detection.}
Similarly, we apply our 2D branch's feature extractor to 2D detector DETR \cite{detr} by replacing its vanilla transformer backbone. We conduct experiments on both pre-trained and scratch backbones and report our performance on the ScanNetV2 2D detection dataset. As shown in Tab.~\ref{tab:2ddection}, our model significantly improves the performance of DETR, demonstrating the strong generalization ability on 2D tasks of our model.

\textbf{Few shot image classification.}
We conduct few-shot image classification experiments on three different benchmarks to explore the feature-extracting ability of PiMAE's image encoder. To verify the effectiveness of PiMAE, we use no extra design for the classifier by only adding a linear layer to the feature encoder, predicting the class based on \textit{[CLS]} token as input.
Tab.~\ref{tab:fewshot} summarizes our results. We see significant improvements from PiMAE pre-training compared to models trained from scratch. Moreover, our performance surpasses previous SOTA self-supervised multi-modal learning method, CrossPoint\cite{crosspoint}.

\begin{table*}[!t]
  \centering
  \small 
  \begin{tabularx}{0.69\textwidth}{lcccccc}
    \toprule
    & \multicolumn{2}{c}{CIFAR-FS 5-way} & \multicolumn{2}{c}{FC100 5-way} & \multicolumn{2}{c}{miniImageNet 5-way} \\
    \cmidrule(lr){2-3} \cmidrule(lr){4-5} \cmidrule(lr){6-7}
    Method  & 1-shot & 5-shot & 1-shot & 5-shot & 1-shot & 5-shot \\
    \midrule
    MAML\cite{MAML} & 58.9 & 71.5 & - & - & 48.7 & 63.1 \\
    Matching Networks\cite{miniImageNet} & - & - & - & - & 43.6 & 55.3 \\
    Prototypical Network\cite{protypical} & 55.5 & 72.0 & 35.3 & 48.6 & 49.3 & 68.2 \\
    Relation Network\cite{relationnetwork} & 55.0 & 69.3 & - & - & 50.4 & 65.3 \\
    \midrule
    CrossPoint\cite{crosspoint}  & 64.5 & 80.1 & - & - & - & - \\
    \midrule
    PiMAE From Scratch  & 62.4 & 76.6 & 37.3 & 50.5 & 50.1 & 66.7 \\
    PiMAE Pre-trained  & \textbf{66.9} & \textbf{80.7} & \textbf{39.0} & \textbf{53.3} & \textbf{55.3} & \textbf{70.2} \\
    \bottomrule
  \end{tabularx}
  \vspace{-0.05in}
   \caption{\textbf{Few-shot image classification on CIFAR-FS, FC100 and miniImageNet test sets.} We report top-1 classification accuracy under 5-way 1-shot and 5-way 5-shot settings. Results of CrossPoint and previous methods are from \cite{crosspoint, cifar-fs}.}
    \vspace{-0.05in}
 \label{tab:fewshot}
\end{table*}

\subsection{Ablation Study}
In this section, we investigate our methods, evaluating the quality of different PiMAE pre-training strategies both qualitatively and quantitatively. 
First, we attempt different alignment strategies between the masked tokens. Next, we pre-train with different reconstruction targets. Then, we ablate performance pre-trained with only a single branch. Finally, we examine our data efficiency by training on limited data. Additional ablation studies on model architecture and masking ratios are in our Appendix. All experiments are based on 3DETR and performed on SUN RGB-D unless otherwise stated.

\begin{figure} [h]
    \centering 
    \includegraphics[width=0.48\textwidth]{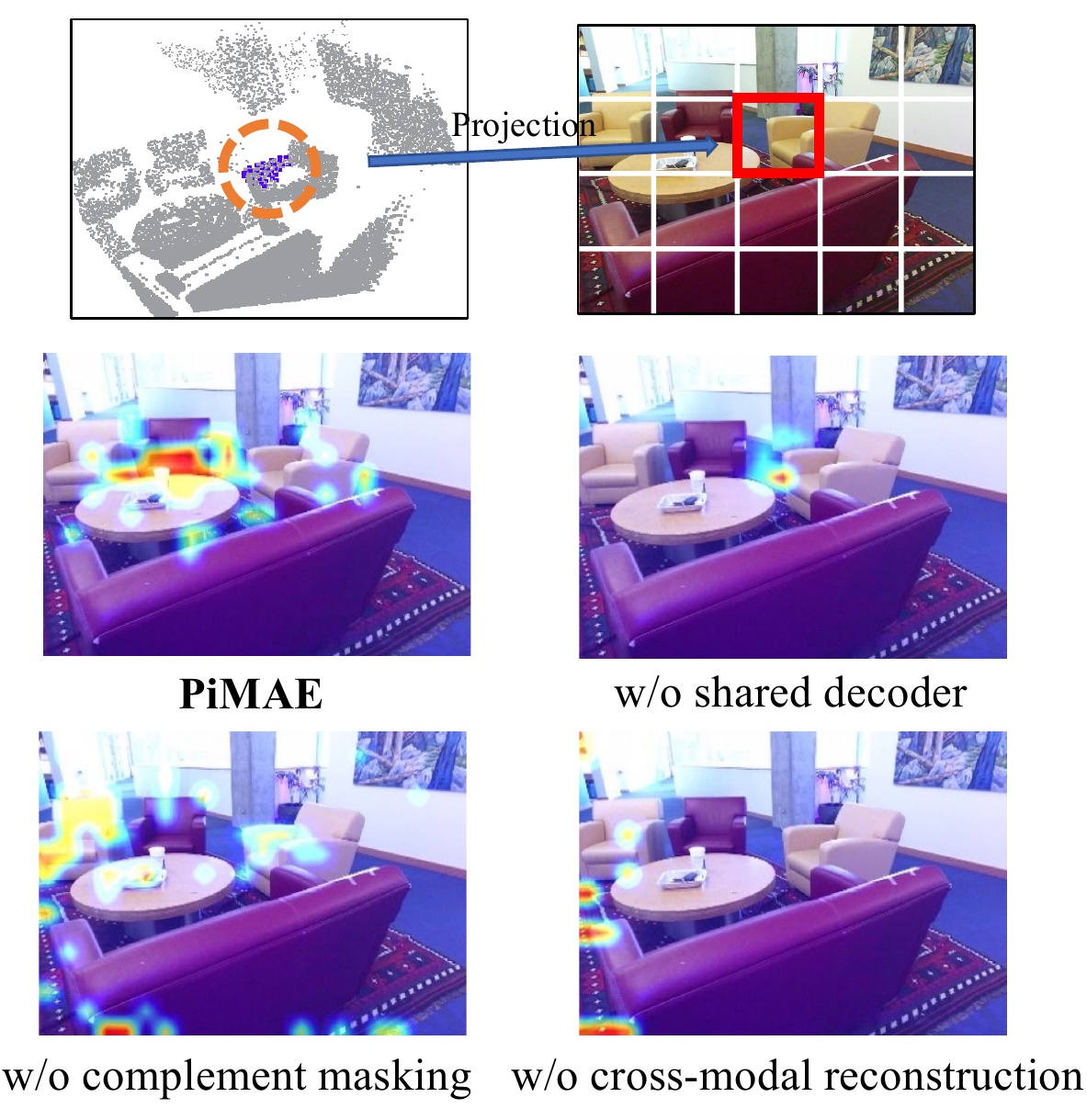} 
    \vspace{-0.15in}
      \vspace{-0.05in}

    \caption{\textbf{Visualization of attention.} The encoder attention between the two modalities is visualized by computing self-attention from the query points (orange circle) to all the visible image tokens. We show the corresponding location (red square) of the query points after projection. See more examples in the Appendix.} 
    \label{Fig:attention} 
    \vspace{-0.05in}
\end{figure}

\textbf{Cross-modal masking.}  
To better study the mask relationships between the two modalities, we design two masking strategies based on projection alignment: uniform masking and complement masking.
Whereas the former masks both modalities in the same pattern that masked portions of one modality will correspond to the other when projected onto it, the latter is the opposite, i.e. a visible point cloud patch will be masked when projected on the image.

We pre-train both the uniform, complement as well as random masking strategies and evaluate their fine-tuning performance on 3D detection. Random masking acts as a baseline where the masking pattern of image and point patches are individually and randomly sampled. As shown in Tab.~\ref{tab:strategy}, masking the tokens from different modalities complementarily gains higher performance than randomly.

Compared to random masking, complement masking enables more cross-modal interactions between patches with diverse semantic information, thus helping our model to transfer 2D knowledge into the 3D feature extractor. However, with the uniform masking strategy, the extracted point cloud features and image features are semantically aligned, so the interaction does not help the model utilize 2D information better. 

Note that in Tab.~\ref{tab:strategy}, we purposely pre-train our model without the cross-modal reconstruction module. This is because the uniform masking strategy projects onto the masked image features, which are semantically weaker and may negatively influence our ablation study.

\begin{table}
  \centering
  \small
  \begin{tabular}{ccc} 
    \toprule
     Masking Strategy & $AP_{25}$ & $AP_{50}$ \\
    \midrule
    Random & 58.0 & 32.9 \\
     Uniform  & 58.1 & 32.6 \\ 
     Complement & \textbf{59.0} & \textbf{33.0} \\
    \bottomrule
  \end{tabular}
    \vspace{-0.05in}
  \caption{\textbf{Comparisons of cross-modality masking strategies.}}
  \label{tab:strategy}
    \vspace{-0.05in}

\end{table}

\textbf{Effect of Cross-modal Reconstruction}. 
Other than reconstructing the inputs, we promote cross-modal reconstruction by demanding point cloud features to reconstruct features or pixels of the corresponding image. We assess the significance of such design by ablating results on 3D detection. Shown in Tab.~\ref{tab:reconstruction}, the additional feature-level cross-modal reconstruction target brings additional performance gains.%
The promoted cross-modal reconstruction at the feature level encourages further interactions between modalities and encodes 2D knowledge into our feature extractor, improving model performance on downstream tasks.

\begin{table}
  \centering
  \small\begin{tabular}{@{}cccccc@{}} 
    \toprule
     \multicolumn{3}{c}{Point Cloud} & \multicolumn{1}{c}{RGB} & \multirow{2}{*}{$AP_{25}$} & \multirow{2}{*}{$AP_{50}$} \\
     \cmidrule(lr){1-3} \cmidrule(lr){4-4} 
    3D Geo & 2D feat & 2D pix & 2D pix \\
    \midrule
     \checkmark &  & & \checkmark & 59.0 & 33.0 \\
     \checkmark &  & \checkmark & \checkmark & 58.0 & 31.6\\
     \checkmark & \checkmark & & \checkmark & \textbf{59.4} & \textbf{33.2}\\
    \bottomrule
  \end{tabular}
\vspace{-0.05in}
   \caption{\textbf{Ablation studies of cross-modal reconstruction targets.} Geo, feat, and pix refer to coordinates, features, and pixels, respectively.}
  \label{tab:reconstruction}
\end{table}

\textbf{Necessity of joint pre-training.} 
To demonstrate the effectiveness and the cruciality of double-branch pre-training in PiMAE, we provide the comparison of PiMAE pre-trained with one branch only. As shown in Tab.~\ref{tab:single-branch}. A critical performance drop can be seen among double-branch PiMAE and single-branch PiMAE. Such evidence reveals the fact that PiMAE learns 3D and 2D features jointly and the cross-modal interactions that we propose help the model to utilize information from both modalities.

Furthermore, to demonstrate the effectiveness of PiMAE's cross-modal interaction design, we visualize the attention map in our shared encoder in Fig.~\ref{Fig:attention}. With our proposed design, PiMAE focuses on more foreground objects with higher attention values, showing a strong cross-modal understanding. See more examples in our Appendix.

\begin{table}
  \centering
  \small
  \scalebox{0.9}{
  \begin{tabular}{@{}lcccc@{}}
    \toprule
     & \multicolumn{2}{c}{3D Object Detection} & \multicolumn{2}{c}{Few-shot image classification} \\
     \cmidrule(lr){2-3} \cmidrule(lr){4-5} 
     Input & $AP_{25}$ & $AP_{50}$ & 5-way 1-shot & 5-way 5-shot \\
    \midrule
    RGB & - & - & 66.3 & 79.5 \\
    Geo & 58.4 & 32.3 & - & - \\
    \midrule
    RGB+Geo & \textbf{59.4} & \textbf{33.2} & \textbf{66.9} & \textbf{80.7} \\
    \bottomrule
  \end{tabular}}
    \vspace{-0.05in}

   \caption{\textbf{Improvement from joint pre-training.} We compare results on 3D object detection (on SUN RGB-D) and few-shot image classification (on CIFAR-FS) tasks when pre-trained with a single-branch PiMAE.}
  \label{tab:single-branch}
\end{table}

\begin{figure} 
    \centering 
    \includegraphics[width=0.4\textwidth]{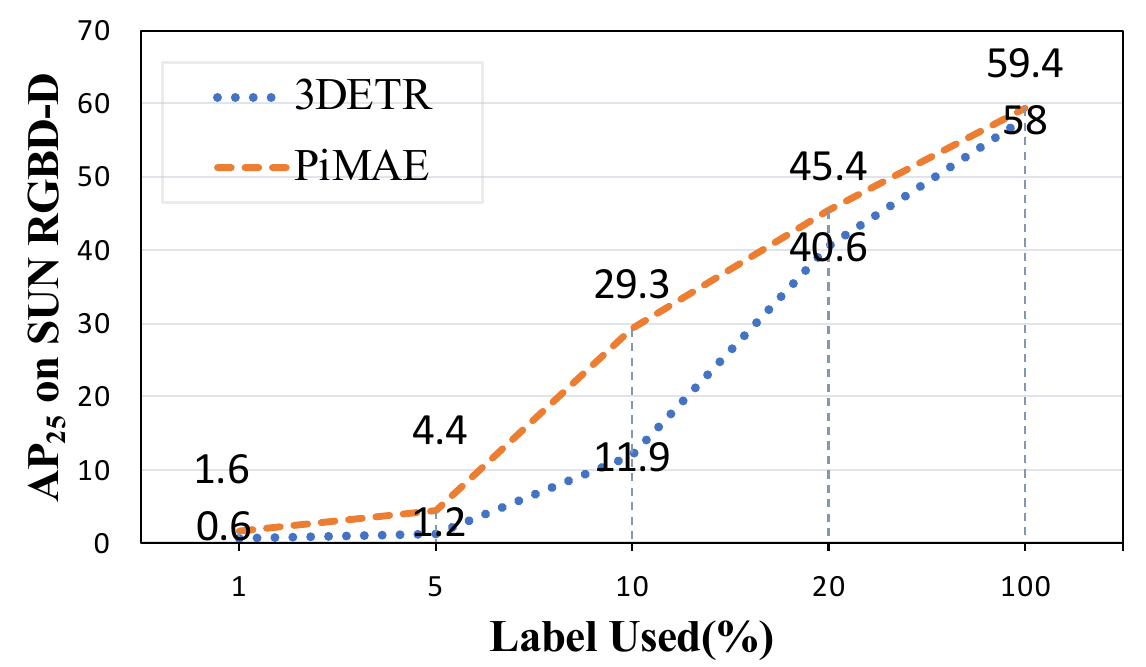} 
    \vspace{-0.05in}
    \caption{\textbf{Illustration of Data Efficiency.} Compared to 3DETR, PiMAE is able to ease the burden of data labeling and increase performance significantly.} 
    \label{Fig:data} 
    \vspace{-0.05in}
\end{figure}

\textbf{Data efficiency.}
We train 3DETR and PiMAE-based 3DETR using limited annotated labels (varying from 1\% to 100\%) while testing on the full val set on SUN RGB-D. As shown in Tab.~\ref{Fig:data}, PiMAE is able to outperform the baseline in every scenario, with the largest difference being 17.4\% ($AP_{25}$) when 10\% of labels are used.

\section{Conclusion}

\label{sec:conc}

In this work, we demonstrate PiMAE's simple framework is an effective and highly interactive multi-modal learning pipeline with strong feature extraction abilities on point cloud and image. 
We design three aspects for promoting cross-modality interaction. First, we explicitly align the mask patterns of both point cloud and image for better feature fusion. Next, we design a shared-decoder to accommodate mask tokens of both modalities. Finally, our cross-modality reconstruction enhances the learned semantics. In our extensive experiments and ablation studies performed on datasets of both modalities, we discover that PiMAE has great potential, improving multiple baselines and tasks.

{\small
\bibliographystyle{ieee_fullname}
\bibliography{PaperForReview}

\begin{thebibliography}{10}\itemsep=-1pt

\bibitem{crosspoint}
Mohamed Afham, Isuru Dissanayake, Dinithi Dissanayake, Amaya Dharmasiri,
  Kanchana Thilakarathna, and Ranga Rodrigo.
\newblock Crosspoint: Self-supervised cross-modal contrastive learning for 3d
  point cloud understanding.
\newblock In {\em Proceedings of the IEEE/CVF Conference on Computer Vision and
  Pattern Recognition (CVPR)}, pages 9902--9912, June 2022.

\bibitem{bachmann2022multimae}
Roman Bachmann, David Mizrahi, Andrei Atanov, and Amir Zamir.
\newblock {MultiMAE}: Multi-modal multi-task masked autoencoders.
\newblock {\em arXiv preprint arXiv:2204.01678}, 2022.

\bibitem{beit}
Hangbo Bao, Li Dong, and Furu Wei.
\newblock {BEiT}: {BERT} pre-training of image transformers.
\newblock 2021.

\bibitem{cifar-fs}
Luca Bertinetto, Joao~F. Henriques, Philip Torr, and Andrea Vedaldi.
\newblock Meta-learning with differentiable closed-form solvers.
\newblock In {\em International Conference on Learning Representations}, 2019.

\bibitem{detr}
Nicolas Carion, Francisco Massa, Gabriel Synnaeve, Nicolas Usunier, Alexander
  Kirillov, and Sergey Zagoruyko.
\newblock End-to-end object detection with transformers.
\newblock {\em CoRR}, abs/2005.12872, 2020.

\bibitem{chen2020simple}
Ting Chen, Simon Kornblith, Mohammad Norouzi, and Geoffrey Hinton.
\newblock A simple framework for contrastive learning of visual
  representations.
\newblock {\em arXiv preprint arXiv:2002.05709}, 2020.

\bibitem{chen2021mocov3}
Xinlei Chen*, Saining Xie*, and Kaiming He.
\newblock An empirical study of training self-supervised vision transformers.
\newblock {\em arXiv preprint arXiv:2104.02057}, 2021.

\bibitem{NEURIPS2020_63c3ddcc}
Ching-Yao Chuang, Joshua Robinson, Yen-Chen Lin, Antonio Torralba, and Stefanie
  Jegelka.
\newblock Debiased contrastive learning.
\newblock In H. Larochelle, M. Ranzato, R. Hadsell, M.F. Balcan, and H. Lin,
  editors, {\em Advances in Neural Information Processing Systems}, volume~33,
  pages 8765--8775. Curran Associates, Inc., 2020.

\bibitem{dai2017scannet}
Angela Dai, Angel~X. Chang, Manolis Savva, Maciej Halber, Thomas Funkhouser,
  and Matthias Nie{\ss}ner.
\newblock Scannet: Richly-annotated 3d reconstructions of indoor scenes.
\newblock In {\em Proc. Computer Vision and Pattern Recognition (CVPR), IEEE},
  2017.

\bibitem{vit}
Alexey Dosovitskiy, Lucas Beyer, Alexander Kolesnikov, Dirk Weissenborn,
  Xiaohua Zhai, Thomas Unterthiner, Mostafa Dehghani, Matthias Minderer, Georg
  Heigold, Sylvain Gelly, Jakob Uszkoreit, and Neil Houlsby.
\newblock An image is worth 16x16 words: Transformers for image recognition at
  scale.
\newblock {\em ICLR}, 2021.

\bibitem{chamferdistance}
Haoqiang Fan, Hao Su, and Leonidas Guibas.
\newblock A point set generation network for 3d object reconstruction from a
  single image, 2016.

\bibitem{MAML}
Chelsea Finn, Pieter Abbeel, and Sergey Levine.
\newblock Model-agnostic meta-learning for fast adaptation of deep networks,
  2017.

\bibitem{fu2022pos}
Kexue Fu, Peng Gao, ShaoLei Liu, Renrui Zhang, Yu Qiao, and Manning Wang.
\newblock Pos-bert: Point cloud one-stage bert pre-training.
\newblock {\em arXiv preprint arXiv:2204.00989}, 2022.

\bibitem{convmae}
Peng Gao, Teli Ma, Hongsheng Li, Jifeng Dai, and Yu Qiao.
\newblock Convmae: Masked convolution meets masked autoencoders.
\newblock {\em arXiv preprint arXiv:2205.03892}, 2022.

\bibitem{gaomimic}
Peng Gao, Renrui Zhang, Hongyang Li, Hongsheng Li, and Yu Qiao.
\newblock Mimic before reconstruct: Enhance masked autoencoders with feature
  mimicking.

\bibitem{kitti}
Andreas Geiger, Philip Lenz, and Raquel Urtasun.
\newblock Are we ready for autonomous driving? the kitti vision benchmark
  suite.
\newblock In {\em Conference on Computer Vision and Pattern Recognition
  (CVPR)}, 2012.

\bibitem{M3AE}
Xinyang Geng, Hao Liu, Lisa Lee, Dale Schuurmans, Sergey Levine, and Pieter
  Abbeel.
\newblock Multimodal masked autoencoders learn transferable representations.
\newblock 2022.

\bibitem{guo2022calip}
Ziyu Guo, Renrui Zhang, Longtian Qiu, Xianzheng Ma, Xupeng Miao, Xuming He, and
  Bin Cui.
\newblock Calip: Zero-shot enhancement of clip with parameter-free attention.
\newblock {\em arXiv preprint arXiv:2209.14169}, 2022.

\bibitem{gwak2020gsdn}
JunYoung Gwak, Christopher~B Choy, and Silvio Savarese.
\newblock Generative sparse detection networks for 3d single-shot object
  detection.
\newblock In {\em European conference on computer vision}, 2020.

\bibitem{he2022masked}
Kaiming He, Xinlei Chen, Saining Xie, Yanghao Li, Piotr Doll{\'a}r, and Ross
  Girshick.
\newblock Masked autoencoders are scalable vision learners.
\newblock In {\em Proceedings of the IEEE/CVF Conference on Computer Vision and
  Pattern Recognition}, pages 16000--16009, 2022.

\bibitem{hess2022masked}
Georg Hess, Johan Jaxing, Elias Svensson, David Hagerman, Christoffer
  Petersson, and Lennart Svensson.
\newblock Masked autoencoders for self-supervised learning on automotive point
  clouds.
\newblock {\em arXiv preprint arXiv:2207.00531}, 2022.

\bibitem{hong2021unbiased}
Youngkyu Hong and Eunho Yang.
\newblock Unbiased classification through bias-contrastive and bias-balanced
  learning.
\newblock {\em Advances in Neural Information Processing Systems},
  34:26449--26461, 2021.

\bibitem{3D-SIS}
Ji Hou, Angela Dai, and Matthias Nießner.
\newblock 3d-sis: 3d semantic instance segmentation of rgb-d scans, 2018.

\bibitem{houji}
Ji Hou, Benjamin Graham, Matthias Nießner, and Saining Xie.
\newblock Exploring data-efficient 3d scene understanding with contrastive
  scene contexts, 2020.

\bibitem{pri3d}
Ji Hou, Saining Xie, Benjamin Graham, Angela Dai, and Matthias Nießner.
\newblock Pri3d: Can 3d priors help 2d representation learning?, 2021.

\bibitem{huang2022tig}
Peixiang Huang, Li Liu, Renrui Zhang, Song Zhang, Xinli Xu, Baichao Wang, and
  Guoyi Liu.
\newblock Tig-bev: Multi-view bev 3d object detection via target inner-geometry
  learning.
\newblock {\em arXiv preprint arXiv:2212.13979}, 2022.

\bibitem{jaritz2022cross}
Maximilian Jaritz, Tuan-Hung Vu, Raoul De~Charette, {\'E}milie Wirbel, and
  Patrick P{\'e}rez.
\newblock Cross-modal learning for domain adaptation in 3d semantic
  segmentation.
\newblock {\em IEEE Transactions on Pattern Analysis and Machine Intelligence},
  2022.

\bibitem{adamw}
Diederik~P. Kingma and Jimmy Ba.
\newblock Adam: A method for stochastic optimization, 2014.

\bibitem{conv1}
Jason Ku, Melissa Mozifian, Jungwook Lee, Ali Harakeh, and Steven Waslander.
\newblock Joint 3d proposal generation and object detection from view
  aggregation, 2017.

\bibitem{2D-driven}
Jean Lahoud and Bernard Ghanem.
\newblock 2d-driven 3d object detection in rgb-d images.
\newblock In {\em 2017 IEEE International Conference on Computer Vision
  (ICCV)}, pages 4632--4640, 2017.

\bibitem{AIST}
Ruilong Li, Shan Yang, David~A. Ross, and Angjoo Kanazawa.
\newblock Ai choreographer: Music conditioned 3d dance generation with aist++,
  2021.

\bibitem{li2022simipu}
Zhenyu Li, Zehui Chen, Ang Li, Liangji Fang, Qinhong Jiang, Xianming Liu,
  Junjun Jiang, Bolei Zhou, and Hang Zhao.
\newblock Simipu: Simple 2d image and 3d point cloud unsupervised pre-training
  for spatial-aware visual representations.
\newblock In {\em Proceedings of the AAAI Conference on Artificial
  Intelligence}, volume~36, pages 1500--1508, 2022.

\bibitem{conv2}
Ming Liang, Bin Yang, Shenlong Wang, and Raquel Urtasun.
\newblock Deep continuous fusion for multi-sensor 3d object detection, 2020.

\bibitem{P4Constrast}
Yunze Liu, Li Yi, Shanghang Zhang, Qingnan Fan, Thomas~A. Funkhouser, and Hao
  Dong.
\newblock P4contrast: Contrastive learning with pairs of point-pixel pairs for
  {RGB-D} scene understanding, 2020.

\bibitem{groupfree}
Ze Liu, Zheng Zhang, Yue Cao, Han Hu, and Xin Tong.
\newblock Group-free 3d object detection via transformers.
\newblock {\em arXiv preprint arXiv:2104.00678}, 2021.

\bibitem{Liu2020TANetR3}
Zhe Liu, Xin Zhao, Tengteng Huang, Ruolan Hu, Yu Zhou, and Xiang Bai.
\newblock Tanet: Robust 3d object detection from point clouds with triple
  attention.
\newblock {\em ArXiv}, abs/1912.05163, 2020.

\bibitem{mao2021dual}
Mingyuan Mao, Renrui Zhang, Honghui Zheng, Teli Ma, Yan Peng, Errui Ding,
  Baochang Zhang, Shumin Han, et~al.
\newblock Dual-stream network for visual recognition.
\newblock {\em Advances in Neural Information Processing Systems},
  34:25346--25358, 2021.

\bibitem{min2022voxel}
Chen Min, Dawei Zhao, Liang Xiao, Yiming Nie, and Bin Dai.
\newblock Voxel-mae: Masked autoencoders for pre-training large-scale point
  clouds.
\newblock {\em arXiv preprint arXiv:2206.09900}, 2022.

\bibitem{misra2021-3detr}
Ishan Misra, Rohit Girdhar, and Armand Joulin.
\newblock {An End-to-End Transformer Model for 3D Object Detection}.
\newblock In {\em {ICCV}}, 2021.

\bibitem{FC100}
Boris~N. Oreshkin, Pau Rodriguez, and Alexandre Lacoste.
\newblock Tadam: Task dependent adaptive metric for improved few-shot learning.
\newblock 2018.

\bibitem{pointmae}
Yatian Pang, Wenxiao Wang, Francis~EH Tay, Wei Liu, Yonghong Tian, and Li Yuan.
\newblock Masked autoencoders for point cloud self-supervised learning.
\newblock {\em arXiv preprint arXiv:2203.06604}, 2022.

\bibitem{peng2021sparse}
Duo Peng, Yinjie Lei, Wen Li, Pingping Zhang, and Yulan Guo.
\newblock Sparse-to-dense feature matching: Intra and inter domain cross-modal
  learning in domain adaptation for 3d semantic segmentation.
\newblock In {\em Proceedings of the IEEE/CVF International Conference on
  Computer Vision}, pages 7108--7117, 2021.

\bibitem{votenet}
Charles~R. Qi, Or Litany, Kaiming He, and Leonidas~J. Guibas.
\newblock Deep hough voting for 3d object detection in point clouds, 2019.

\bibitem{F-PointNet}
Charles~R. Qi, Wei Liu, Chenxia Wu, Hao Su, and Leonidas~J. Guibas.
\newblock Frustum pointnets for 3d object detection from rgb-d data, 2017.

\bibitem{qi2016pointnet}
Charles~R Qi, Hao Su, Kaichun Mo, and Leonidas~J Guibas.
\newblock Pointnet: Deep learning on point sets for 3d classification and
  segmentation.
\newblock {\em arXiv preprint arXiv:1612.00593}, 2016.

\bibitem{rukhovich2021fcaf3d}
Danila Rukhovich, Anna Vorontsova, and Anton Konushin.
\newblock Fcaf3d: Fully convolutional anchor-free 3d object detection.
\newblock {\em arXiv preprint arXiv:2112.00322}, 2021.

\bibitem{imvoxelnet}
Danila Rukhovich, Anna Vorontsova, and Anton Konushin.
\newblock Imvoxelnet: Image to voxels projection for monocular and multi-view
  general-purpose 3d object detection.
\newblock {\em CoRR}, abs/2106.01178, 2021.

\bibitem{shi2020pv}
Shaoshuai Shi, Chaoxu Guo, Li Jiang, Zhe Wang, Jianping Shi, Xiaogang Wang, and
  Hongsheng Li.
\newblock Pv-rcnn: Point-voxel feature set abstraction for 3d object detection.
\newblock In {\em Proceedings of the IEEE/CVF Conference on Computer Vision and
  Pattern Recognition}, pages 10529--10538, 2020.

\bibitem{shi20193d}
S Shi, X Wang, H~Pointrcnn Li, et~al.
\newblock 3d object proposal generation and detection from point cloud.
\newblock In {\em Proceedings of the IEEE conference on computer vision and
  pattern recognition, Long Beach, CA, USA}, pages 15--20, 2019.

\bibitem{protypical}
Jake Snell, Kevin Swersky, and Richard~S. Zemel.
\newblock Prototypical networks for few-shot learning, 2017.

\bibitem{sunrgbd}
Shuran Song, Samuel~P. Lichtenberg, and Jianxiong Xiao.
\newblock Sun rgb-d: A rgb-d scene understanding benchmark suite.
\newblock In {\em 2015 IEEE Conference on Computer Vision and Pattern
  Recognition (CVPR)}, pages 567--576, 2015.

\bibitem{DSS}
Shuran Song and Jianxiong Xiao.
\newblock Deep sliding shapes for amodal 3d object detection in rgb-d images,
  2015.

\bibitem{relationnetwork}
Flood Sung, Yongxin Yang, Li Zhang, Tao Xiang, Philip H.~S. Torr, and
  Timothy~M. Hospedales.
\newblock Learning to compare: Relation network for few-shot learning, 2017.

\bibitem{transformers}
Ashish Vaswani, Noam Shazeer, Niki Parmar, Jakob Uszkoreit, Llion Jones,
  Aidan~N. Gomez, Lukasz Kaiser, and Illia Polosukhin.
\newblock Attention is all you need, 2017.

\bibitem{miniImageNet}
Oriol Vinyals, Charles Blundell, Timothy Lillicrap, koray kavukcuoglu, and Daan
  Wierstra.
\newblock Matching networks for one shot learning.
\newblock In D. Lee, M. Sugiyama, U. Luxburg, I. Guyon, and R. Garnett,
  editors, {\em Advances in Neural Information Processing Systems}, volume~29.
  Curran Associates, Inc., 2016.

\bibitem{wang2021pointaugmenting}
Chunwei Wang, Chao Ma, Ming Zhu, and Xiaokang Yang.
\newblock Pointaugmenting: Cross-modal augmentation for 3d object detection.
\newblock In {\em Proceedings of the IEEE/CVF Conference on Computer Vision and
  Pattern Recognition}, pages 11794--11803, 2021.

\bibitem{wu2022eda}
Yanmin Wu, Xinhua Cheng, Renrui Zhang, Zesen Cheng, and Jian Zhang.
\newblock Eda: Explicit text-decoupling and dense alignment for 3d visual and
  language learning.
\newblock {\em arXiv preprint arXiv:2209.14941}, 2022.

\bibitem{xie2020pointcontrast}
Saining Xie, Jiatao Gu, Demi Guo, Charles~R Qi, Leonidas Guibas, and Or Litany.
\newblock Pointcontrast: Unsupervised pre-training for 3d point cloud
  understanding.
\newblock In {\em European conference on computer vision}, pages 574--591.
  Springer, 2020.

\bibitem{xie2021simmim}
Zhenda Xie, Zheng Zhang, Yue Cao, Yutong Lin, Jianmin Bao, Zhuliang Yao, Qi
  Dai, and Han Hu.
\newblock Simmim: A simple framework for masked image modeling.
\newblock In {\em International Conference on Computer Vision and Pattern
  Recognition (CVPR)}, 2022.

\bibitem{Xu_2018_CVPR}
Bin Xu and Zhenzhong Chen.
\newblock Multi-level fusion based 3d object detection from monocular images.
\newblock In {\em Proceedings of the IEEE Conference on Computer Vision and
  Pattern Recognition (CVPR)}, June 2018.

\bibitem{PointFusion}
Danfei Xu, Dragomir Anguelov, and Ashesh Jain.
\newblock Pointfusion: Deep sensor fusion for 3d bounding box estimation, 2017.

\bibitem{yang2022boosting}
Hao Yang, Chen Shi, Yihong Chen, and Liwei Wang.
\newblock Boosting 3d object detection via object-focused image fusion.
\newblock {\em arXiv preprint arXiv:2207.10589}, 2022.

\bibitem{zhang2022i-mae}
Kevin Zhang and Zhiqiang Shen.
\newblock i-mae: Are latent representations in masked autoencoders linearly
  separable?
\newblock {\em arXiv preprint arXiv:2210.11470}, 2022.

\bibitem{pointm2ae}
Renrui Zhang, Ziyu Guo, Peng Gao, Rongyao Fang, Bin Zhao, Dong Wang, Yu Qiao,
  and Hongsheng Li.
\newblock Point-m2ae: Multi-scale masked autoencoders for hierarchical point
  cloud pre-training.
\newblock {\em arXiv preprint arXiv:2205.14401}, 2022.

\bibitem{zhang2022pointclip}
Renrui Zhang, Ziyu Guo, Wei Zhang, Kunchang Li, Xupeng Miao, Bin Cui, Yu Qiao,
  Peng Gao, and Hongsheng Li.
\newblock Pointclip: Point cloud understanding by clip.
\newblock In {\em Proceedings of the IEEE/CVF Conference on Computer Vision and
  Pattern Recognition}, pages 8552--8562, 2022.

\bibitem{zhang2022monodetr}
Renrui Zhang, Han Qiu, Tai Wang, Xuanzhuo Xu, Ziyu Guo, Yu Qiao, Peng Gao, and
  Hongsheng Li.
\newblock Monodetr: Depth-aware transformer for monocular 3d object detection.
\newblock {\em arXiv preprint arXiv:2203.13310}, 2022.

\bibitem{zhang2022learning}
Renrui Zhang, Liuhui Wang, Yu Qiao, Peng Gao, and Hongsheng Li.
\newblock Learning 3d representations from 2d pre-trained models via
  image-to-point masked autoencoders.
\newblock {\em arXiv preprint arXiv:2212.06785}, 2022.

\bibitem{zhang2023parameter}
Renrui Zhang, Liuhui Wang, Yali Wang, Peng Gao, Hongsheng Li, and Jianbo Shi.
\newblock Parameter is not all you need: Starting from non-parametric networks
  for 3d point cloud analysis.
\newblock In {\em Proceedings of the IEEE/CVF Conference on Computer Vision and
  Pattern Recognition}, 2023.

\bibitem{zhang2022can}
Renrui Zhang, Ziyao Zeng, Ziyu Guo, and Yafeng Li.
\newblock Can language understand depth?
\newblock In {\em Proceedings of the 30th ACM International Conference on
  Multimedia}, pages 6868--6874, 2022.

\bibitem{zhang2021self}
Zaiwei Zhang, Rohit Girdhar, Armand Joulin, and Ishan Misra.
\newblock Self-supervised pretraining of 3d features on any point-cloud.
\newblock In {\em Proceedings of the IEEE/CVF International Conference on
  Computer Vision}, pages 10252--10263, 2021.

\bibitem{H3DNet}
Zaiwei Zhang, Bo Sun, Haitao Yang, and Qixing Huang.
\newblock H3dnet: 3d object detection using hybrid geometric primitives, 2020.

\bibitem{pointTrans}
Hengshuang Zhao, Li Jiang, Jiaya Jia, Philip H.~S. Torr, and Vladlen Koltun.
\newblock Point transformer.
\newblock {\em CoRR}, abs/2012.09164, 2020.

\bibitem{zhao2021graph}
Han Zhao, Xu Yang, Zhenru Wang, Erkun Yang, and Cheng Deng.
\newblock Graph debiased contrastive learning with joint representation
  clustering.
\newblock In {\em IJCAI}, pages 3434--3440, 2021.

\bibitem{zheng2020end}
Minghang Zheng, Peng Gao, Renrui Zhang, Kunchang Li, Xiaogang Wang, Hongsheng
  Li, and Hao Dong.
\newblock End-to-end object detection with adaptive clustering transformer.
\newblock {\em arXiv preprint arXiv:2011.09315}, 2020.

\bibitem{voxelnet}
Yin Zhou and Oncel Tuzel.
\newblock Voxelnet: End-to-end learning for point cloud based 3d object
  detection, 2017.

\bibitem{zhu2022pointclip}
Xiangyang Zhu, Renrui Zhang, Bowei He, Ziyao Zeng, Shanghang Zhang, and Peng
  Gao.
\newblock Pointclip v2: Adapting clip for powerful 3d open-world learning.
\newblock {\em arXiv preprint arXiv:2211.11682}, 2022.

\end{thebibliography}
}

\clearpage
\renewcommand\thesection{\Alph{section}}

\section*{\Large Appendix}

\setcounter{section}{0}

\section{Supplementary Implementation Details}
\label{sec:implm}

In this section, we first introduce the descriptions of dataset we verified the effects of our pre-training scheme on. Then, we elaborate on the details of the baseline models we used in our empirical studies. Finally, we detail the specific configurations used in all of our experiments.

\subsection{Datasets and metric}
\label{sec:dataset}
SUN RGB-D\cite{sunrgbd} is a challenging large-scale 3D indoor dataset, consisting of 10,335 RGB-D images with labeled 3D bounding boxes for 37 categories. Depth images are converted to point clouds using provided camera poses, and we follow the standard 5285, 5050 splits for the training and testing stages, respectively. We report the accuracy on the test set of SUN RGB-D using the mean Average Precision at two different IoU thresholds, 0.25 and 0.5, respectively.

ScanNetV2 \cite{dai2017scannet}
is a 3D interior scene dataset with rich annotations, consisting of 1513 indoor scenes and 18 object classes. The labels include semantic labels, per-point instances, 2D and 3D bounding boxes. For 3D object detection, we use the common metrics for evaluation \cite{votenet}, measuring the mean Average Precision ($mAP$) under two IoU thresholds of 0.25 and 0.5. For 2D detection, we follow \cite{detr} to report the Average Precision ($AP$) under 0.5, 0.75, and 1.0.

KITTI \cite{kitti} is a widely adopted outdoor 3D object detection benchmark, consisting of 7481 training images and 7518 test images. To evaluate our approach, we follow \cite{zhang2022monodetr} and adopt the Average Precision ($AP$) of 3D bounding boxes under three-level difficulties, easy, moderate and hard. Detection scores for the car category under the intersection-over-union (IoU) threshold of 0.7 are reported.

CIFAR-FS\cite{cifar-fs}, FC100\cite{FC100}, miniImageNet\cite{miniImageNet}, are widely used few-shot image classification datasets, and we use these datasets to evaluate PiMAE's image feature extractor. The top-1 accuracy under 5-way 1-shot and 5-way 5-shot settings on different datasets is adopted as the evaluation metric.      

\subsection{Baseline Approaches}
We evaluate our interactive multi-modal training pre-training scheme by fine-tuning three state-of-the-art 3D object detectors, and our baseline implementations rigorously follow their publicly released codes.

\textbf{3DETR\cite{misra2021-3detr}} is a simple, end-to-end 3D detection pipeline that does not require finely crafted 3D detection backbone. Instead, its versatile attention-based backbone maximally preserves the vanilla Transformer blocks to reach comparable performance with CNN-based detectors.

\textbf{Group-Free 3D\cite{groupfree}}
\noindent is another approach implementing the Transformer models on 3D object detection task, using both a well-designed query locations for objects and an ensembling of detection results. Unlike PointNet-based networks \cite{qi2016pointnet, F-PointNet, shi20193d} that create a local grouping scheme for each object candidate, Group-Free uses an attention mechanism on all the point cloud points. 

\textbf{DETR\cite{detr}}
\noindent is an end-to-end 2D object detection that uses a Transformer architecture to force unique predictions with bipartite matching. DETR can quickly and efficiently make predictions of the relations between objects and the global image context from a small set of object queries.

\textbf{MonoDETR\cite{zhang2022monodetr}}
\noindent is a novel, state-of-the-art DETR-based model for monocular 3D detection that does not rely on depth supervision, anchors, or Non-Maximum Suppression (NMS). It modifies DETR's vanilla transformer to incorporate depth estimates and predicts 3D annotations from the depth information inherent in images.

\begin{table} [t]
\centering
\begin{tabular}{l|c}
\toprule
Config & Value \\
\midrule
optimizer & AdamW\cite{adamw} \\
base lr & 1e-3 \\
weight decay & 0.05 \\
batch size & 256 \\
lr schedule & cosine decay \\
warmup epochs & 15 \\
epoch & 400 \\
augmentation & None \\
\bottomrule
\end{tabular}
\vspace{-0.05in}
\caption{\textbf{Pre-training configuration.} }
\label{tab:pretrain}
\end{table}

\subsection{Pre-training Details}
\label{sec:pretraindetails}
The encoder and decoder architectures in PiMAE follow the standard ViT\cite{vit} design, which consists of several Transformer blocks. In our PiMAE, the number of Transformer blocks for specific encoders and shared-decoders is both set to 3, while the numbers for specific decoders and shared-decoders are set to 2 and 1, respectively. For encoders, each Transformer block has 256 hidden dimensions and 4 heads for the multi-head self-attention module. For decoders, the numbers are adjusted to 192 and 3.

For the point cloud branch, we sample 2048 points from each 3D scene in SUN RGB-D\cite{sunrgbd}, following previous work~\cite{pointm2ae}. For the image branch, we adjust the resolution of each image to $256 \times 352$, and we use a patch size of 16 to patchify images. The specific configuration for pre-training PiMAE is given in Tab.~\ref{tab:pretrain}.

\subsection{Fine-tuning Protocol on SUN RGB-D, ScanNetV2, and KITTI}
\label{sec:finetunedetails}
For fine-tuning on GroupFree3D~\cite{groupfree}, we insert our 3D feature extractor into the pipeline. TODO:inserted where? 
Compared to the original configuration, the only modification here is tuning the learning rate on the encoder lower to $lr={3e-5}$ to preserve the pre-trained prior. For detection on ScanNetV2, we lower the encoder learning rate to $lr={6e-5}$.

For experiments with 3DETR~\cite{misra2021-3detr}, our encoder consists of six Transformer blocks pre-trained with PiMAE. The exact setups as the original~\cite{misra2021-3detr} are then used for fine-tuning, except that we apply a reduced learning rate of $lr={1e-5}$ to the encoder.

For DETR~\cite{detr}, we replace the vanilla Transformer encoder with our image branch feature extractor and our joint-encoder pre-trained on SUN RGB-D. The depth of the encoder is unchanged and we only apply a reduced learning rate of $lr={1e-5}$ to the encoder. We perform 2D object detection on the ScannetV2 2D detection dataset.

For MonoDETR\cite{zhang2022monodetr}, we replace its depth encoder with our 3-layer 3D feature extractor and follow the original configuration for training.

\section{Additional Ablation Study}
\label{sec:ablation}
In this section, we give more ablation studies for further analysis of PiMAE. 

\textbf{Ablation study on Pipeline Architecture.} 
During the reconstruction stage, as proposed in Sec.~\ref{sec:decoder}, a shared decoder architecture is adopted. The encoded features are first disentangled by the cross-modality decoder, and reconstructions are completed afterward with task-specific decoders. 
From Tab.~\ref{tab:architecture}, 
we find the additional shared-decoder design performance-enhancing, 
as it considers the cross-modal influence of masked tokens. Specifically, shared-decoder is a novel contribution of PiMAE, and we find it non-trivial, because the interactions in the masked tokens improve feature extraction.

\textbf{Ablation Study on Masking Ratio.}
As reported in Tab.~\ref{tab:mask_ratio}, we examined several masking ratios for PiMAE and find that the the model learns the best latent features when the masking ratio is set to 60\%. 

\section{Additional Visualization}
\label{sec:vis}

\begin{figure*} [t]
  \centering
  \begin{subfigure}{\linewidth}
  \centering
    {\includegraphics[width=0.9\linewidth]{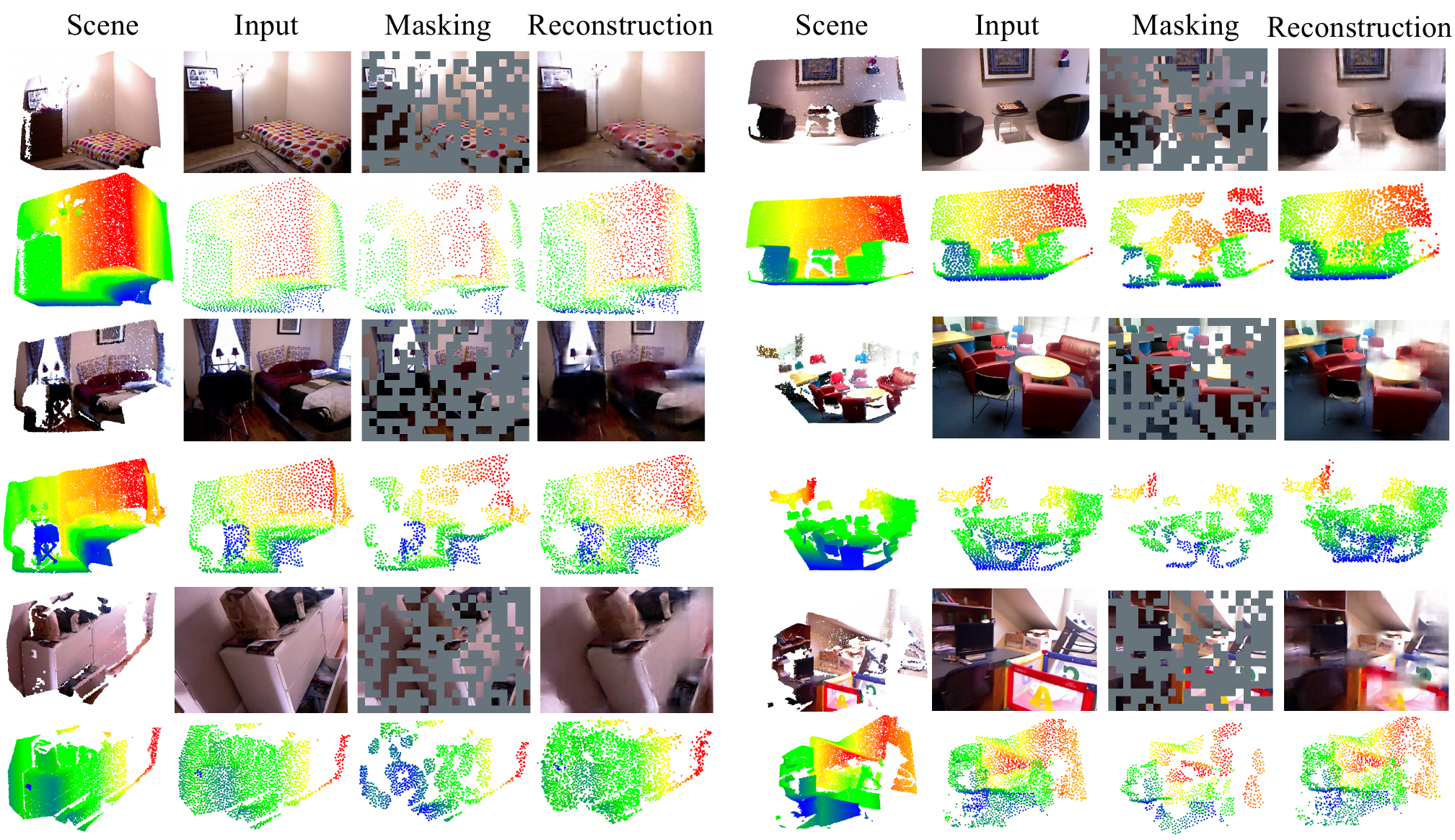}
    }
  \end{subfigure}
  \hfill
  \vspace{-0.05in}
  \caption{\textbf{Visualization of reconstruction results.} Our model is trained with 60\% masking ratio. Point clouds are colored for better visualization purpose. PiMAE generalizes well for different scenes and reconstructs masked images (odd rows) and masked point clouds (even rows) simultaneously.}
  \label{fig:reconstruction1}
\end{figure*}

\textbf{Reconstruction Results.}
In Fig.~\ref{fig:reconstruction1}. We provide more examples of reconstruction visualizations. PiMAE simultaneously reconstructs masked point clouds and images with clear reconstructions reflecting semantic understanding.

\begin{figure*} 
  \centering
\begin{subfigure}{\linewidth}
  \centering
    {\includegraphics[width=0.95\linewidth]{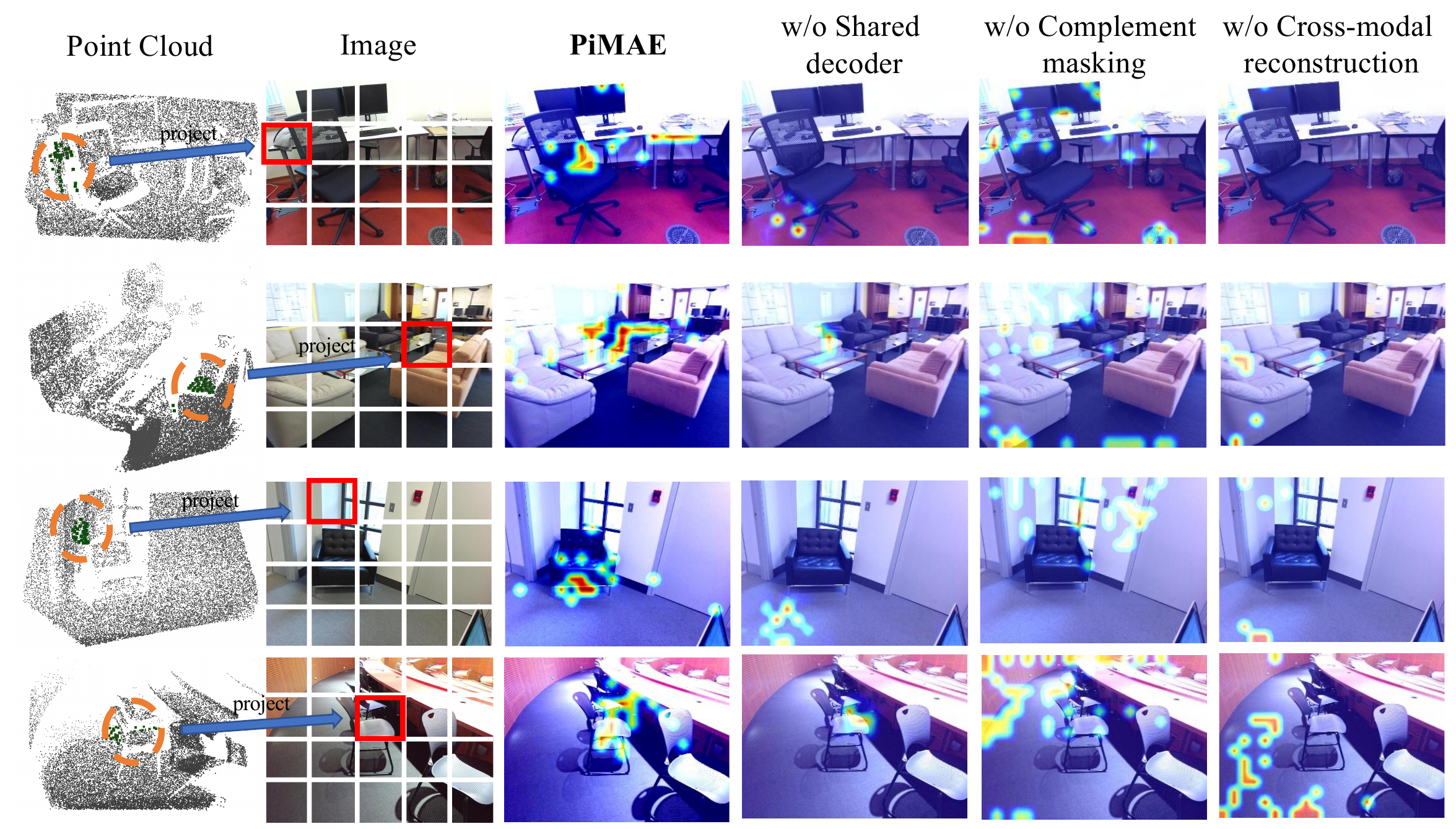}
    }
  \end{subfigure}
  \hfill
  \caption{\textbf{Visualization of encoder attention, point cloud as query.} The encoder's attention between two modalities is visualized by computing self-attention from the query points (orange circle) to all the visible image tokens. Highest values are shown in red. We show the corresponding location (red square) of the query points after projection. From left to right shows ablation of PiMAE with different designs, including our final proposal, and settings that exclude shared-decoder, complement masking strategy and cross-modal reconstruction, respectively. }
  \vspace{0.8cm}
  \label{fig:attn_pc2img}

  \begin{subfigure}{\linewidth}
  \centering
    {\includegraphics[width=0.95\linewidth]{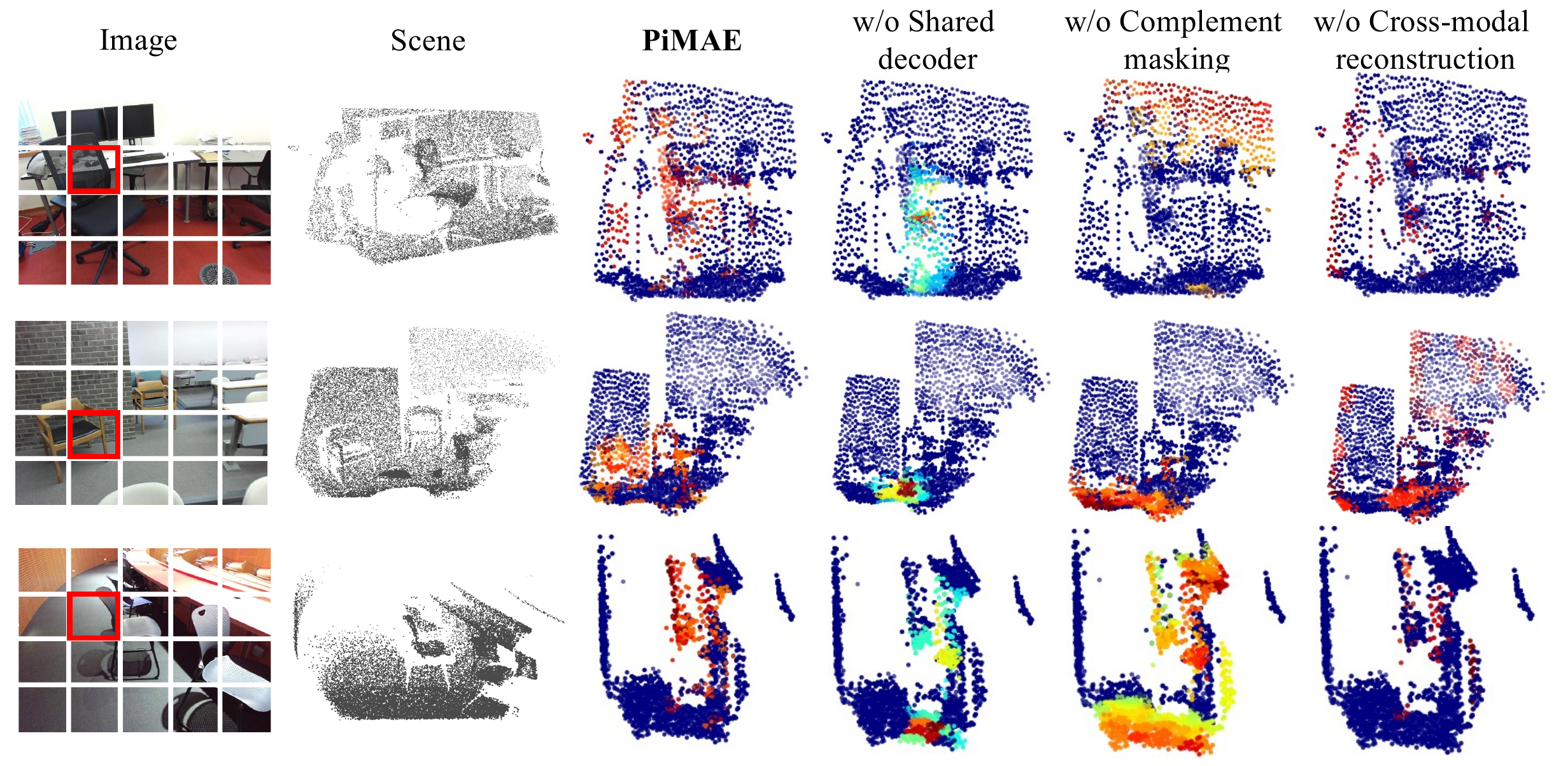}
    }
  \end{subfigure}
  \hfill
  \caption{\textbf{Visualization of encoder attention, image as query.} The encoder attention between the two modalities is visualized by computing self-attention from the query of an image token (red square) to all the point cloud tokens. Highest values are shown in red. The attention intensity in the point cloud corresponds with the image patch query, showing the effectiveness of our cross-modal interactions during pre-training. }
  \label{fig:attn_img2pc}

    \vspace{0.5in}

\end{figure*}

\textbf{Activation of Feature Map.}
This section provides more attention map examples generated by PiMAE's shared-encoder, where features from the two modalities first interact explicitly. By examining the self-attention weights, we can gain better insights on PiMAE's multi-modal interactions.
We compute the self-attention from a point cloud token to all image tokens and show the attention values. In Fig.~\ref{fig:attn_pc2img}, PiMAE is able to attend to more foreground objects and with higher attention values, while other designs either attend to unrelated backgrounds(e.g. row \textcolor{blue}{3}, col \textcolor{blue}{4}), or have rather low attention values (e.g. row \textcolor{blue}{2}, col \textcolor{blue}{4}).

We also compute the self-attention from a image token to all point cloud tokens and display the attention weights. As shown in Fig.~\ref{fig:attn_img2pc}, given a image token as query, PiMAE accurately attends to the corresponding objects in the point cloud with highest values, showing a strong understanding of both 2D and 3D features.

\begin{figure*} [t]
  \centering
  \begin{subfigure}{0.98\linewidth}
  \centering
    {\includegraphics[width=\linewidth]{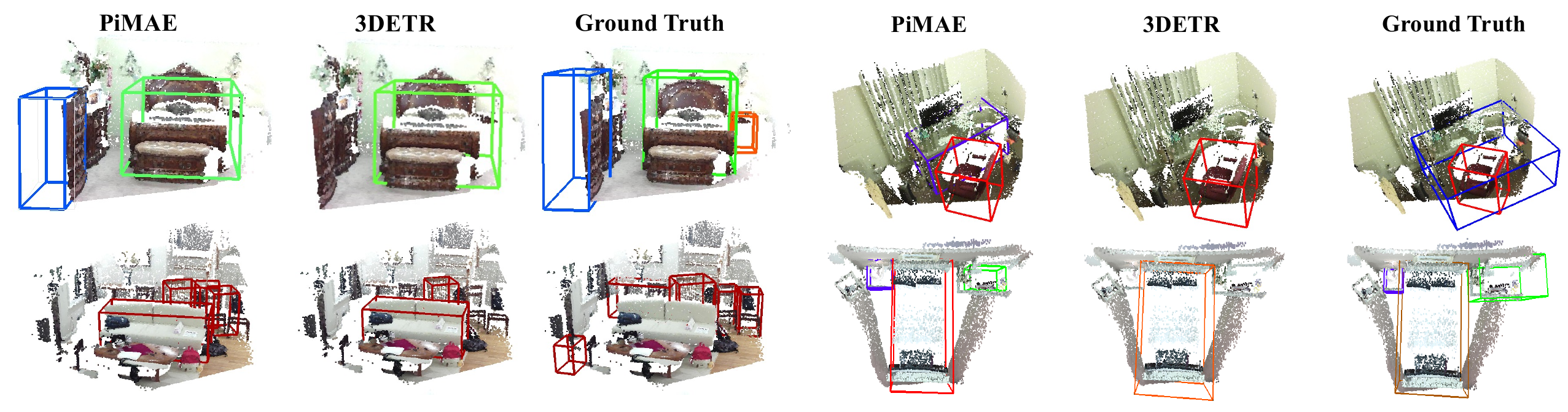}
    }
  \end{subfigure}
  \hfill
  \vspace{-0.1in}
  \caption{\textbf{Visualization of predictions on SUN RGB-D validation set.} We correctly detect more objects.}
  \label{fig:detection}
\end{figure*}

\textbf{Object Detection.}
We display more qualitative results comparing PiMAE and baseline in Fig.~\ref{fig:detection}. On top of 3DETR~\cite{misra2021-3detr}, with PiMAE pre-training, we are able to detect more objects with more precise boxes.

\begin{table}
  \centering
  \small\begin{tabular}{@{}cccc@{}} 
    \toprule
     Encoder & Decoder & $AP_{25}$ & $AP_{50}$ \\
    \midrule
     3+3 & 0+3 & 58.0 & 30.2 \\
     3+3 & 1+2 & \textbf{59.4} & \textbf{33.2} \\
     \midrule
     3+3 & 1+3 & 58.1 & 32.8 \\
     2+2 & 1+2 & 57.5 & 30.8 \\
    \bottomrule
  \end{tabular}
    \vspace{-0.05in}

  \caption{\textbf{Ablation studies on model architecture.} a+b in encoder denotes specific encoders of a-layers ViT, and shared-encoder of b-layers ViT. c+d in decoder denotes c-layers ViT for shared-decoder and d-layers ViT for specific decoders. Experiments are based on 3DETR and performed on SUN RGB-D.} 
  \label{tab:architecture}
    \vspace{-0.05in}
\end{table}

\begin{table} [t]
\centering
\begin{tabular}{c|cccc}
\toprule
Mask Ratio & $AP_{25}$ & $AP_{50}$\\
\midrule
 50\% & 58.7 & 33.1 \\
 60\% & \textbf{59.4} & \textbf{33.2} \\
 70\% & 58.4 & 33.0 \\
 80\% & 57.5 & 32.4 \\
\bottomrule
\end{tabular}
\vspace{-0.05in}
\caption{\textbf{Ablation study on masking ratios.} Experiments with different masking ratio are conducted, and we report detection accuracy based of 3DETR on SUN RGB-D val set.}
\label{tab:mask_ratio}
\end{table}

\clearpage

\begin{table*}[h]
\centering
\small
 \resizebox{0.95\linewidth}{!}{
\begin{tabularx}{1.02\linewidth}{l|cccccccccc|cc}
\toprule
Methods & bed & table & sofa & chair & toilet & desk & dresser & nightstd & bookshf & bathtub & $AP_{25}$ & $AP_{50}$ \\
\midrule
DSS\cite{DSS} & 78.8 & 50.3 & 53.5 & 61.2 & 78.9 & 20.5 & 6.4 & 15.4 & 11.9 & 44.2 & 42.1 & - \\
2D-driven\cite{2D-driven} & 64.5 & 37.0 & 50.4 & 48.3 & 80.4 & 27.9 & 25.9 & 41.9 & 31.4 & 43.5 & 45.1 & - \\
PointFusion\cite{PointFusion} & 68.6 & 31.0 & 53.8 & 55.1 & 83.8 & 17.2 & 23.9 & 32.3 & \textbf{37.7} & 37.3 & 45.4 & - \\
F-PointNet\cite{F-PointNet} & 81.1 & 51.1 & 61.1 & 64.2 & 90.9 & 24.7 & 32.0 & 58.1 & 33.3 & 43.3 & 54.0 & - \\
VoteNet\cite{votenet} & 83.0 & 47.3 & 64.0 & 75.3 & 90.1 & 22.0 & 29.8 & 62.2 & 28.8 & 74.4 & 57.7 & - \\
\midrule
 3DETR\cite{misra2021-3detr} & 81.8 & 50.0 & 58.3 & 68.0 & 90.3 & 28.7 & 28.6 & 56.6 & 27.5 & 77.6 & 58.0 & 30.3 \\
 +Ours & 85.4 & 48.9 & 62.5 & 69.0 & 93.8 & 28.2 & 33.0 & 62.8 & 30.4 & 80.3 & 59.4(+1.4) & 33.2(+2.9) \ \ \ \ \ \ \\
\midrule
GroupFree3D\cite{groupfree} & \textbf{87.8} & 53.8 & 70.0 & \textbf{79.4} & 91.1 & \textbf{32.6} & 36.0 & 66.7 & 32.5 & 80.0 & 63.0 & 45.2 \\
+Ours & 85.4 & \textbf{55.1} & \textbf{73.3} & 78.1 & \textbf{96.0} & 31.5 & \textbf{40.8} & \textbf{67.8} & 28.4 & \textbf{89.1} & \textbf{64.6}(+1.6) & \textbf{46.2}(+1.0) \ \ \ \ \ \ \\
\bottomrule
\end{tabularx}
}
\vspace{-0.05in}
\caption{\textbf{3D objection detection results on SUN RGB-D validation set.} Single-class precision is reported with $AP_{25}$. Results of previous methods are taken from \cite{votenet,misra2021-3detr,groupfree}.}
\label{tab:overall}
\vspace{-0.05in}
\end{table*}

\clearpage

\end{document}